\documentclass{article}
\usepackage{PRIMEarxiv}
\usepackage{amsmath, amssymb, graphicx, url, algorithm2e}
\usepackage{times}
\usepackage{xr-hyper} 
\usepackage{hyperref}
\usepackage{subfiles}
\usepackage{thmtools}
\usepackage{xcolor}
\usepackage{caption}
\usepackage{subfigure}
\usepackage{graphicx}
\usepackage{float}
\usepackage[rightcaption]{sidecap}
\usepackage{tikz}
\usepackage{comment}

\usepackage[toc,page]{appendix}
\usetikzlibrary{shapes.geometric,positioning,bayesnet,calc,intersections,through,backgrounds}
\usetikzlibrary{arrows,shapes,plotmarks,positioning}
\usetikzlibrary{decorations.markings}
\usepackage{tkz-euclide}
\usepackage{pgfplots}
 \pgfplotsset{compat=1.17}
\tikzset{>=stealth'}
\tikzstyle{graphnode} =
   [circle,draw=black,minimum size=22pt,text centered,text
     width=22pt,inner sep=0pt]
\tikzstyle{var}   =[graphnode,fill=white]
\tikzstyle{vardashed}   =[graphnode,draw=gray,fill=white]
\tikzstyle{obs}   =[graphnode,fill=black,text=white]
\tikzstyle{obsgrey}   =[graphnode,draw=white,fill=lightgray,text=black]
\tikzstyle{par}    =[graphnode,draw=white,fill=red,text=black]
 \tikzstyle{crucial} =[graphnode,draw=white,fill=yellow,text=black]
\tikzstyle{fac}   =[rectangle,draw=black,fill=black!25,minimum size=5pt]
\tikzstyle{facprior} =[rectangle,draw=black,fill=black,text=white,minimum size=5pt]
\tikzstyle{edge}  =[draw=white,double=black,very thick,-]
\tikzstyle{blueedge}  =[draw=white,double=blue,very thick,-]
\tikzstyle{rededge}  =[draw=white,double=red,very thick,-]
\tikzstyle{prior} =[rectangle, draw=black, fill=black, minimum size=
5pt, inner sep=0pt]
\tikzstyle{dirprior} = [circle, draw=black, fill=black, minimum
size=5pt, inner sep=0pt]
\tikzstyle{dot_node}=[draw=black,fill=black,shape=circle]
\usepackage{enumitem}
\setlist[itemize]{leftmargin=*,itemsep=0em,topsep=0em}
\setlist[enumerate]{leftmargin=*,itemsep=0em,topsep=0em}
\usepackage{csquotes}

\usepackage{soul}
\usepackage{tcolorbox}
\usepackage[super]{nth}
\usepackage{nccmath}
\usetikzlibrary{patterns}
\usepackage[abs]{overpic}
\usepackage{diagbox}

\newtheorem{lemma}{Lemma}
\newtheorem{definition}{Definition}
\newtheorem{proposition}{Proposition}
\newtheorem{theorem}{Theorem}
\newtheorem{example}{Example}

\newcommand{\cX}{{\cal X}}
\newcommand{\cY}{{\cal Y}}
\newcommand{\cZ}{{\cal Z}} 
\newcommand{\cW}{{\cal W}}

\usepackage{thmtools}
\usepackage{xcolor}
\usepackage{graphicx}
\usepackage{booktabs}
\usepackage{caption}
\usepackage{tikz}
\usepackage{comment}
\usepackage[toc,page]{appendix}
\usetikzlibrary{shapes.geometric,positioning,bayesnet,calc,intersections,through,backgrounds}
\usetikzlibrary{arrows,shapes,plotmarks,positioning}
\usetikzlibrary{decorations.markings}
\usepackage{tkz-euclide}
\usepackage{pgfplots}
\tikzset{>=stealth'}
\tikzstyle{graphnode} =
   [circle,draw=black,minimum size=22pt,text centered,text
     width=22pt,inner sep=0pt]
\tikzstyle{var}   =[graphnode,fill=white]
\tikzstyle{vardashed}   =[graphnode,draw=gray,fill=white]
\tikzstyle{obs}   =[graphnode,fill=black,text=white]
\tikzstyle{obsgrey}   =[graphnode,draw=white,fill=lightgray,text=black]
\tikzstyle{par}    =[graphnode,draw=white,fill=red,text=black]
 \tikzstyle{crucial} =[graphnode,draw=white,fill=yellow,text=black]
\tikzstyle{fac}   =[rectangle,draw=black,fill=black!25,minimum size=5pt]
\tikzstyle{facprior} =[rectangle,draw=black,fill=black,text=white,minimum size=5pt]
\tikzstyle{edge}  =[draw=white,double=black,very thick,-]
\tikzstyle{blueedge}  =[draw=white,double=blue,very thick,-]
\tikzstyle{rededge}  =[draw=white,double=red,very thick,-]
\tikzstyle{prior} =[rectangle, draw=black, fill=black, minimum size=
5pt, inner sep=0pt]
\tikzstyle{dirprior} = [circle, draw=black, fill=black, minimum
size=5pt, inner sep=0pt]
\tikzstyle{dot_node}=[draw=black,fill=black,shape=circle]
\usepackage{enumitem}
\setlist[itemize]{leftmargin=*,itemsep=0em,topsep=0em}
\setlist[enumerate]{leftmargin=*,itemsep=0em,topsep=0em}
\usepackage{csquotes}
\usepackage{soul}
\usepackage{tcolorbox}
\usepackage[super]{nth}
\usepackage{nccmath}
\usetikzlibrary{patterns}
\usepackage[abs]{overpic}
\usepackage{diagbox}
\usepackage{nicematrix}
\usepackage{url}

\newcommand\independent{\protect\mathpalette{\protect\independenT}{\perp}}
\def\independenT#1#2{\mathrel{\rlap{$#1#2$}\mkern2mu{#1#2}}}
\title{Bounding Probabilities Of Causation \\Through The Causal Marginal Problem}

\author{
  Numair Sani  \\
  \texttt{snumair1@jh.edu} \\
   \And
  Atalanti A. Mastakouri \\
  \texttt{atalanti@amazon.de} \\
     \And
  Dominik Janzing\\
  \texttt{janzind@amazon.de}
}

\begin{document}

\maketitle

\begin{abstract}
 
Probabilities of Causation play a fundamental role in decision making in law, health care and public policy. Nevertheless, their point identification is challenging, requiring strong assumptions such as monotonicity. In the absence of such assumptions, existing work requires multiple observations of datasets that contain the {\it same} treatment and outcome variables, in order to establish bounds on these probabilities.
However, in many clinical trials and public policy evaluation cases, there exist independent datasets that examine the effect of a {\it different} treatment each on the same outcome variable. Here, we outline how to significantly tighten existing bounds on the probabilities of causation, by imposing counterfactual consistency between SCMs constructed from such independent datasets (`causal marginal problem'). Next, we describe a new information theoretic approach on falsification of counterfactual probabilities, using conditional mutual information to quantify counterfactual influence. The latter generalises to arbitrary discrete variables and number of treatments, and renders the causal marginal problem more interpretable. Since the question of `tight enough' is left to the user, we provide an additional method of inference when the bounds are unsatisfactory: 
A maximum entropy based method that defines a metric for the space of plausible SCMs and proposes the entropy maximising SCM for inferring counterfactuals in the absence of more information. %Second, a Bayesian inference framework that allows for the construction of credible intervals and confirmation measures for the Probabilities of Causation.

\end{abstract}

\section{Introduction}
Probabilities of Causation defined in \cite{pearl2009causality} represent a class of causal quantities that have applications in law, public policy and health-care \cite{mueller2022personalized, faigman2014group}. 
While causal parameters such as the Average Treatment Effect are \emph{point identified} from the observed data distribution under reasonable assumptions such as exogeneity, Probabilities of Causation are not. 
Specifically, in addition to exogeneity, these require monotonicity to hold for point identification \cite{pearl2022probabilities, khoury1989measurement}.
 However, when these assumptions are not justified, and, as such, \emph{point identification} is not possible, the goal is redefined as \emph{partial identification}, i.e. bounding these probabilities as much as possible using observational data.

There exists a rich literature on partial identification and its use in bounding causal quantities; some examples include \cite{manski1990nonparametric, tian2000probabilities, balke1994counterfactual, robins1989probability, dawid2017probability}. The specific question of bounding Probabilities of Causation has also been explored in \cite{tian2000probabilities, zhang2022partial,cuellar2018causal, robins1989probability, dawid2017probability, padh2022stochastic, duarte2021automated, sachs2022general}, however, our contribution here differs fundamentally from the existing work. We only consider the specific causal graph in Fig. \ref{fig:causal-graph} containing two treatments and a single outcome arranged in a collider structure. Additionally we assume discrete valued treatments and outcome, and instead of having access to the joint distribution over both treatments $X$ and $Y$ and the outcome $Z$, we only have access to marginals of this joint distribution, i.e. $\mathbb{P}(X, Z)$ and $\mathbb{P}(Y, Z)$. While this is a simple example, it captures a common situation occurring in the world; that is many independent studies looking at the same outcome of interest but studying a different treatment. A concrete example of this can be seen in randomised control trials (RCTs), where competing pharmaceutical companies are testing different treatments for the same disease. Then the joint distribution refers to a hypothetical experimental design in which both treatments are independently randomised. Therefore, while \cite{duarte2021automated, zhang2022partial, li2022probabilities, pearl2022probabilities, cuellar2018causal, dawid2017probability} generate bounds on counterfactuals such as the Probabilities of Causation, they all assume access to a joint distribution over all the variables of interest. In contrast, we only assume access to marginals of the joint distribution over variables of interest. This scenario imposes additional challenges since the joint is not in general identified from the marginals, making it challenging to apply existing bounds. 

To address these challenges, we first present an information theoretic perspective on reasoning about counterfactuals when combining marginal distributions, and use conditional mutual information to provide results on falsification of counterfactual probabilities. Next, using the consistency constrains shown in \cite{gresele2022causal}, we significantly tighten existing bounds on Probabilities of Causation. Since `tight enough' is a decision left to the user of our method, we provide one additional method of inference when the bounds are unsatisfactory. This is a Maximum Entropy based method to select the `best' Structural Causal Model (SCM). While learning SCMs from data has been studied in \cite{hyttinen2012learning, shimizu2011directlingam}, in our setting the SCM is not identified from the marginals of the observed data. In this setting, we apply the Maximum Entropy Principle \cite{jaynes1988relation} to learn a SCM consistent with the data and use its entropy as a measure of evidence. Causal versions of Maximum Entropy have been discussed in \cite{janzing2021causal} and applied in \cite{mejia2022obtaining} to combine datasets, but they have not been used before to infer SCMs from data. 

Our work bears close similarity to \cite{zeitler2022causal}, however here we outline the key differences: Since we consider a particular causal graph structure, we are able to provide additional results by symbolically solving the relevant linear programs. Additionally, while existing works find bounds for the treatment effect, we are able to provide symbolic bounds that tighten existing bounds on all of the Probabilities of Causation. Additionally, we provide an information theoretic view on the falsification of SCMs when additional variables are available, which provides a simple interpretation of some of the results of \cite{gresele2022causal}.
Further, we introduce a Maximum Entropy based SCM learning framework to pick the SCM that obeys the observed marginal constraints while maximising entropy. %Finally, our Bayesian framework allows for the \emph{confirmation} of values of counterfactuals, as opposed to their falsification - which plays an important role when we have access to limited randomised trials on the same outcome variable of interest.

The remaining of this paper is organised as follows. In Section \ref{sec:preliminaries} we present an overview of counterfactuals and probabilities of causation, and in \ref{sec:causal-marginal} we provide an overview of the Causal Marginal problem \cite{gresele2022causal} as well as rigorously establish the relationship between falsifying SCMs and restricting the bounds on Probabilities of Causation. Then, we provide an information theoretic approach to counterfactual inference, by falsifying SCMs given additional variables in Section \ref{sec:information_theoretic_counterfactuals}. Since information theoretic bounds for marginal problems are usually not tight \cite{ChavesLMGJS2014}, we use a linear programming based approach to establish theoretical conditions under which we can tighten the bounds on the Probabilities of Causation. We show how, under which assumptions, and by how much we can tighten the existing bounds in Section \ref{sec:bounding_poc}. In the event that the bounds are not satisfactorily tight, we provide a Maximum Entropy based method for counterfactual inference in Section \ref{sec:imprecise_probabilities}. Finally, we provide closing remarks in Section \ref{sec:discussion}. All proofs are provided in the Appendix.

\section{Preliminaries}\label{sec:preliminaries}
In this section we briefly introduce the relevant concepts utilised by our paper, as well as provide references for in depth reviews on each of the topic. We start by defining our causal model, which forms the basis of the rest of our causal reasoning. 

While there exist many formulations of causal models in the literature such as the Finest Fully Randomized Causally Interpretable Structured Tree Graph (FFRCISTG) of \cite{robins1986new} and the agnostic causal model of \cite{spirtes2000causation}, we utilise the SCM defined in \cite{pearl2009causality}. Formally, a SCM $\mathcal{M}$ is defined as a tuple $\langle \mathbf{U}, \mathbf{V}, \mathcal{F}, \mathbb{P} \rangle$ where $\mathbf{U}$ and $\mathbf{V}$ represent a set of exogenous and endogenous random variables respectively. $\mathcal{F}$ represents a set of functions that determine the value of $V \in \mathbf{V}$ through $v \leftarrow f_V(pa_V, u_V)$ where $pa_V$ denotes the parents of $V$ and  $u_V$ denotes the values of the noise variables relevant to $V$. $\mathbb{P}$ denotes the joint distribution over the set of noise variables $\mathbf{U}$, and since the noise variables $\mathbf{U}$ are assumed to be mutually independent, the joint distribution $\mathbb{P}(\mathbf{U})$ factorises into the product of the marginals of the individual noise distributions. $\mathcal{M}$ induces an observational data distribution on $\mathbf{V}$, and is associated with a Directed Acyclic Graph (DAG) $\mathcal{G}$. 

Defining an SCM allows us to define submodels, potential responses and counterfactuals, as defined in \cite{pearl2009causality}. Given a causal model $\mathcal{M}$ and a realisation $x$ of random variables $\mathbf{X} \subset \mathbf{V}$, a submodel $\mathcal{M}_x$ corresponds to deleting from $\mathcal{F}$ all functions that set values of elements in $\mathbf{X}$ and replacing them with constant functions $X = x$. The submodel captures the effect of intervention $do(X = x)$ on $\mathcal{M}$. Given a subset $\mathbf{Y} \subset \mathbf{V}$, the potential response $\mathbf{Y}_x(u)$ denotes the values of $Y$ that satisfy $\mathcal{M}_x$ given value $u$ of the exogenous variables $\mathbf{U}$. And so, the counterfactual $\mathbf{Y}_x(u) = y$ represents the scenario where the potential response $\mathbf{Y}_x(u) $ is equal to $y$, if we possibly contract to fact, set $X = x$. When $u$ is generated from $P(\mathbf{U})$, we obtain counterfactual random variables $\mathbf{Y}_{x}$ that have a corresponding probability distribution. 
%\cite{pearl2009causality} defines a three-step procedure for deriving expressions for these counterfactual probabilities. \numair{Should say a bit more about this mysterious three-step procedure}. 
\newcommand{\yshift}{-4em}
\newcommand{\xshift}{5em}
\newcommand{\xzcol}{red!60!white}
\newcommand{\yzcol}{blue!60!white}

\begin{figure}[t]
\centering\subfigure[]
 {\begin{tikzpicture}
 \centering
 \node (X) [obs,thick] {$X$};
 \node (Y) [obs, xshift=2*\xshift] {$Y$};
 \node (Z) [obs, xshift=\xshift, yshift=\yshift] {$Z$};
 \edge [color=\xzcol, thick] {X}{Z};
 \edge [color=\yzcol, thick] {Y}{Z};
 \plate[yshift=-0.em,
 inner sep=0.5em,
 ultra thick,
 color=\yzcol] {M2}{(Y) (Z)}{\textcolor{\yzcol}{\textbf{RCT 2}}};
 \tikzset{plate caption/.style={caption, node distance=0, inner sep=0pt, below left=5pt and 0pt of #1.south,text height=.2em,text depth=0.3em}}
 \plate[yshift=0.3em,
 inner sep=0.5em,
 ultra thick,
 color=\xzcol] {M1}{(X) (Z)}{\textcolor{\xzcol}{\textbf{RCT 1}}}; \end{tikzpicture}\label{fig:causal-graph}}\quad\quad\quad\quad\quad \subfigure[]{\begin{tikzpicture}
 \centering
 \node (X) [obs,thick] {$X$};
 \node (Z) [obs, xshift=2em, yshift=\yshift] {$Z$};
 \node (N_Z) [fac, thick, xshift = 5em, yshift = -2em] {$N_Z$};
 \edge [color=\xzcol, thick] {X}{Z};
 \edge [color=\xzcol, thick] {N_Z}{Z};
 \tikzset{plate caption/.style={caption, node distance=0, inner sep=0pt, below left=5pt and 0pt of #1.south,text height=.2em,text depth=0.3em}}
 \end{tikzpicture}
 \label{fig:teaser}}
\caption{(a) \small \textbf{Causal graph representing independent clinical trials studying identical outcomes.}
 Given two RCTs studying the same outcome $Z$ but with two different treatments $X$ and $Y$, combining these two RCTs restricts the Probabilities of Causation more effectively then either RCT by itself. (b) \small \textbf{Information theoretic view of SCM $X \rightarrow Z$.} Under the response function parameterization, the noise term $N_Z$ can be viewed as the influence of the remaining part of the world on the relationship between $X$ and $Z$ }
\end{figure}
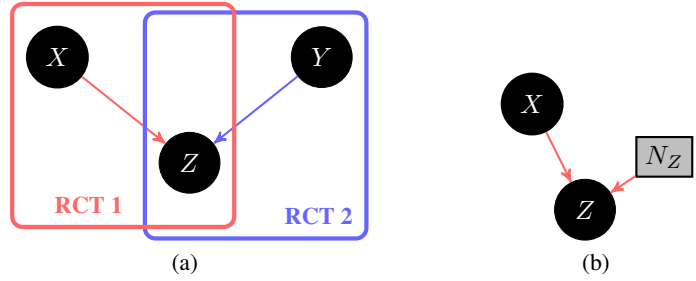

An important class of counterfactual probabilities that have applications in law, medicine, public policy are known as the Probabilities of Causation \cite{pearl2009causality}. These are a set of five counterfactual probabilities related to the Probability of Necessity and Sufficiency (PNS), which we define as follows. Given binary random variables $Z$ and $X$, where $Z$ is the outcome and $X$ is the exposure, let $z$ denote the event that the random variable $Z$ has value $1$ and let $z^{\prime}$ denote $Z$ obtaining value $0$, and a similar notation is followed for $x$ and $x^{\prime}$. Then, the PNS is defined as
\begin{align}
    PNS \equiv \mathbb{P}(Z_x = z, Z_{x^{\prime}} = z^{\prime})
\end{align}
In words, PNS represents the joint probability that the counterfactual random variable $Z_x$ takes on value $z$ and the counterfactual random variable $Z_{x^{\prime}}$ takes on value $z^{\prime}$. Under conditions of exogeneity, defined as $Z_x \independent X$, the rest of the Probabilities of Causation such as Probability of Necessity (PN) and Probability of Sufficiency (PS), are all defined as functions of PNS (see Theorem 9.2.11 in \cite{pearl2009causality}). Consequently, when PNS is identified, all the other Probabilities of Causation are straightforwardly identified from the observed data as well. 

However, to identify PNS, we must make assumptions such as \emph{monotonocity} \cite{tian2000probabilities}, which we may not be justified in making in settings involving experimental drugs, legal matters and occupational health. Although the PNS is no longer point identified, it can still be meaningfully bounded using tools from the partial identification literature. An important bound on PNS is defined in \cite{tian2000probabilities}, and is presented below, with $p_{11}$, $p_{10}$ and $p_{00}$ denoting $\mathbb{P}(Z = 1 \mid X = 1)$, $\mathbb{P}(Z = 1 \mid X = 0)$, and $\mathbb{P}(Z = 0 \mid X = 0)$ respectively. 
\begin{align}
    \max[0, p_{11} - p_{10}] \leq &PNS \leq \min[p_{11}, p_{00}]
\end{align}
Before we can demonstrate how to tighten the bounds on the Probabilities of Causation\footnote{In our setting we only consider the Probabilities of Causation since these have been examined extensively in the literature. Nevertheless, our theory straightforwardly applies to other counterfactual probabilities such as $\mathbb{P}(Z_{x^{\prime}} = z, Z_{x} = z^{\prime})$ and on all functions of it, which we discuss in Section \ref{additional_counterfactuals} of the Appendix.}, we need to introduce additional concepts including the response function parameterisation of a SCM \cite{balke1994counterfactual} (referred to as the canonical parametersation of a SCM in \cite{peters2017elements}) as well as the Causal Marginal problem \cite{gresele2022causal}, which we do in the next section.

\section{Counterfactuals And Causal Marginals}\label{sec:causal-marginal}
In this section we first provide a brief overview of the Causal Marginal problem, which has been studied in \cite{gresele2022causal}. Then we rigorously describe the relationship between falsifying SCMs and bounding the probabilities of causation. For all the following section we assume a causal graph that has a v-structure as in Fig. \ref{fig:teaser}.

\subsection{Causal Marginal Problem}\label{ssec:cmp}
Defining an SCM entails degrees of freedom that are irrelevant even for counterfactual statements (for instance rescaling the noise). It only matters which function $f_Z(.,N_Z)$ is induced by $N_Z$. To remove this irrelevant ambiguity the following `response function' or `canonical' representation \cite{balke1994counterfactual,peters2017elements} is convenient: 

\begin{definition}[response function representation]
An SCM $Z= f_Z(W,N_Z)$ with some $N_Z\independent W$ where $W,Z$ are variables with finite ranges $\cW,\cZ$ has `response function representation' if $N_Z$ attains values in the set of functions $\cW \to \cZ$ and thus $f_Z(W,n_Z) = n_Z(W)$. 
\end{definition}
This way, $P(N_Z)$ is a distribution over {\it functions}.
Now that we examine the binary variable setting, we will have the cases $W=X$ and $W=(X,Y)$ where $N_Z$ attains $4$ or $16$ values, respectively. 
%We start by describing the response function parameterisation for a bivariate SCM over $Z$ and $X$, where $X$ is the treatment and $Z$ is the outcome. The SCM is given as
%\begin{align*}
% X &:= N_X \qquad Z := f(X, N_Z)
%\end{align*}
%Here $N_X$ is a noise variable that generates $X$, and $Z$ is generated by $f$, which takes as inputs realisations of $X$ and $N_Z$, and outputs a value of $Z$. 
In the first case, the $4$ functions read $\left\{\mathbf{0}, \mathbf{1}, \mathbf{ID}, \mathbf{NOT} \right\}$ where the first two are constant functions and $\mathbf{ID}$ and $\mathbf{NOT}$ denote the identity and logical inversion, respectively.
Let $P(N_Z)$ be parameterised as $[a_0, a_1, a_2, a_3]$ where each $a_i$ represents the probability mass placed on the respective functions in $\left\{\mathbf{0}, \mathbf{1}, \mathbf{ID}, \mathbf{NOT} \right\}$.

% Given a value of $X$, $Z$ is a deterministic function of $N_Z$, so $N_Z$ can be thought of as a switch variable over these four functions. This allows us to reparameterise the SCM in terms of the distribution over noise variables $N_X, N_Z$. 

%Under this reparameterisation, $\mathbb{P}(N_X)$ will coincide with the marginal $\mathbb{P}(X)$ of the observational distribution $\mathbb{P}(Z, X)$. However, $\mathbb{P}(N_Z)$, which is a probability distribution over the set $\{\mathbf{0}, \mathbf{1}, \mathbf{ID}, \mathbf{NOT} \}$ needs four constraints in order to uniquely identify it. 

When $W = X$, the observed distribution $\mathbb{P}(X, Z)$ can be used to impose linear constraints on $P(N_Z)$: two constraints can be obtained by enforcing consistency between the SCM and the observed distribution $\mathbb{P}(X, Z)$, and one 
additional constraint can be obtained using the property of probability distributions summing to one. This leaves one free parameter, denoted by $\lambda_X$ that uniquely parameterises the SCM from $X \rightarrow Z$, described below, with a detailed derivation provided in the Appendix Section \ref{lambda_x_derivation}. 
\begin{align}\label{eq:scm-parameterisation}
 \begin{pmatrix}a_0\\ a_1\\ a_2\\ a_3 \end{pmatrix} = 
 \begin{pmatrix} 0 \\ 1 - p_{00} - p_{01} \\ p_{00} \\ p_{01} \end{pmatrix} 
 + \lambda_X \begin{pmatrix}1 \\ 1 \\ -1 \\ -1\end{pmatrix} 
\end{align}
Where $\lambda_X \in [\lambda^{\min}_X, \lambda^{\max}_X]$ and $\lambda^{\min}_X = \max[0, p_{00} + p_{01} - 1]$ and $\lambda^{\max}_X = \min[p_{00}, p_{01}]$. Here $p_{00} \equiv \mathbb{P}(Z = 0 \mid X = 0)$ and $p_{01} \equiv \mathbb{P}(Z = 0 \mid X = 1)$.

\cite{gresele2022causal} showed that given three binary random variables $X$, $Y$ and $Z$, and access only to the marginal distributions $\mathbb{P}(X, Z)$ and $\mathbb{P}(Y, Z)$ with $0 < \mathbb{P}(X) < 1$ and $0 < \mathbb{P}(Y) < 1$, when the causal graph follows the structure in \ref{fig:causal-graph}, further range restrictions on $\lambda_X$ may be obtained. This was achieved by constructing SCMs over $X \rightarrow Z$ and $Y \rightarrow Z$, denoted by $\mathcal{M}_X$ and $\mathcal{M}_Y$ from the observed marginals, and then enforcing \emph{counterfactual consistency} between $\mathcal{M}_X$ and $\mathcal{M}_Y$. Counterfactual consistency is imposed by parameterising the SCM over $\{X, Y\} \rightarrow Z$ -denoted as $\mathcal{M}$- using the response function parameterisation as
\begin{align}
 X := N_X \quad Y := N_Y \quad Z := h_{N_Z}(X, Y)
\end{align}
where $N_Z$ indexes $16$ possible functions $h : \{ X \times Y\} \rightarrow Z$. The projection operator is defined as 

\begin{align}
 \mathcal{P}^{X}_x : h \rightarrow h(x, Y) = f(Y) \qquad
 \mathcal{P}^{Y}_y : h \rightarrow h(X, y) = f(X) 
\end{align}

This projection operator allows us to define the counterfactual consistency constraints between $\mathcal{M}_X$ and $\mathcal{M}_Y$ as
\begin{align}
 a_j(\lambda_X) &= \sum_{y = 0}^1 \mathbb{P}(Y = y) \sum_{k = 0}^{15} \mathbb{I}\left\{\mathcal{P}^{Y}_y(h_k) = f_j(X) \right\}c_k\\
 b_j(\lambda_Y) &= \sum_{x = 0}^1 \mathbb{P}(X = x) \sum_{k = 0}^{15} \mathbb{I}\left\{\mathcal{P}^{X}_x(h_k) = f_j(Y) \right\}c_k\\
\intertext{where $c_k$ denotes $P(N_Z=k)$, the weights on the response functions in $\mathcal{M}$ and $k \in {0, \dots, 15}$, while $a_j$ and $b_j$ denote the weights on the response functions for the SCMs $\mathcal{M}_X$ and $\mathcal{M}_Y$ respectively, and $ j \in \{0, 1, 2, 3\}$. These can be succinctly expressed in matrix form as}
&\qquad\mathbf{a}(\lambda_X) = \mathbf{A}\mathbf{c} \quad \mathbf{b}(\lambda_Y) = \mathbf{B}\mathbf{c}
\end{align}
The definition of $\mathbf{A}$ and $\mathbf{B}$ can be found in \cite{gresele2022causal} Equation 15. The set of values of $\mathbf{c}$ that satisfy the above constraints forms a convex polytope, which  we denote by $\mathcal{C}$. The proof of this statement can be found in \cite{gresele2022causal}, and we describe $\mathcal{C}$ in Section \ref{convex-polytope} of the Appendix. By comparing the range of $\lambda_X$ obtained from $\mathbb{P}(Z, X)$ to the range of $\lambda_X$ that satisfies these equations, we may be further able to restrict the range of $\lambda_X$.
However, while \cite{gresele2022causal} demonstrate an example in which the range of $\lambda_X$ can be restricted, the criteria that $\mathbb{P}(X, Z)$ and $\mathbb{P}(Y, Z)$ must satisfy to restrict this range, as well as its magnitude is not discussed. In this paper, we establish such criteria and consequently use it to tighten the bounds on the probability of causation. To do this, we next establish the connection between $\lambda_X$ and the probabilities of causation. 

\subsection{Relating $\lambda_X$ To Counterfactuals}
In Section \ref{ssec:cmp} we saw the link between SCMs and $\lambda_X$, and how restricting the range of $\lambda_X$ corresponds to falsifying SCMs over $X \rightarrow Z$. As we prove in Section \ref{lambda_x_counterfactual} of the Appendix, $\lambda_X$ also captures the counterfactual influence of a model, i.e. the `unnormalised probability that $Z$ would be different if $X$ is different, given that we observe values of $Z$ and $X$', represented as $p_{00} + p_{01} - 2\lambda_X$. Lower values of $\lambda_X$ correspond to models with stronger values of counterfactual influence. So, probabilities of causation can also be expressed in terms of $\lambda_X$ by following Pearl's three-step process of abduction, action and prediction \cite{pearl2009causality}. Lemma \ref{lem:link_lambda} establishes the link between the Probabilities of Causation and $\lambda_X$, as well as recovers the bounds from \cite{tian2000probabilities} on PNS. 
\begin{lemma}[relating $\lambda_X$ to PNS]\label{lem:link_lambda}
When $X$ and $Z$ are binary random variables and $X \rightarrow Z$, PNS $= p_{00} - \lambda_X$. The bounds on $\lambda_X$ derived from observing $\mathbb{P}(X, Z)$ recover the bounds on PNS from \cite{tian2000probabilities}.
\begin{align}
 \max[0, p_{11} - p_{10}] \leq p_{00} - \lambda_X \leq \min[p_{11}, p_{00}]
\end{align}
\end{lemma}
So, tightening the bounds on the Probabilities of Causation directly corresponds to falsifying SCMs. While imposing consistency constraints on counterfactuals is one way to tighten the bounds on the probabilities of causation (which we explore rigorously in Sec. \ref{sec:bounding_poc}), enumerating the relevant constraints in the case of more than two treatments is cumbersome. As we will show in the next section, an information theoretic approach provides a simple and elegant framework to reason about falsification of SCMs in high dimensions (when a high number treatment variables are provided or when variables can attain many different values), offering additional interpretability insights.

\section{Information Theoretic Counterfactual Inference}\label{sec:information_theoretic_counterfactuals}

In this section we explore the falsification of SCMs from an information theoretic point of view. We limit our discussion to variables with finite range to avoid technical subtleties such as infinite mutual information for deterministic relations. Let $X,Y,Z$ attain values in the finite sets $\cX,\cY,\cZ$. For a family of SCMs over $X \rightarrow Z$ consistent with $\mathbb{P}(X, Z)$, information theoretic measures such as mutual information and entropy provide an interpretable perspective on falsifying SCMs. To demonstrate this, we view the noise $N_Z$ as a random variable capturing (and subsequently coarsening) the effect of the remaining causal ancestors of $Z$ on $Z$.
We briefly recall that for two random variables $U,V$, the conditional Shannon entropy reads 
$
H(U|V):= - \sum_{u,v} \mathbb{P}(U=u,V=v) \log \mathbb{P}(U=u|V=v), 
$ 
and that the conditional mutual information reads $I(U: V \mid W) = H(U|W) + H(V|W) - H(U,V| W)$, see \cite{cover}. 
 %We first define Shannon entropy and mutual information below.
%\numair{Then the mutual information between $N_Z$ and $Z$ captures the counterfactual influence of an SCM}. 
%To formally prove this statement, we first define the Shannon entropy and mutual information for discrete random variables as \cite{cover}
%{\small
%\begin{align*}
%H(Z) &= -\sum_{z} \mathbb{P}(Z = z)\log{\mathbb{P}(Z = z)}\\
%I(X: Z) &= \sum_{x} \sum_{z} \mathbb{P}(Z = z, X = z)\log{\frac{\mathbb{P}(Z = z, X = z)}{\mathbb{P}(Z = z) \mathbb{P}(X = z)}}
%\end{align*}
%}

%\begin{lemma}\label{lem:indcond} 
%For any three random variables $U,V,W$ with $U \independent V$ we have 
%\begin{align*}
%I(U: W) \leq I(U : W \mid V). 
%\end{align*}
%\end{lemma} 

Next, for an observed marginal $\mathbb{P}(X, Z)$ generated via $Z = f_Z(X,N_Z)$, note that $X$ and $N_Z$ fully determine $Z$ ($H(Z \mid X,N_Z) = 0$), so the entropy of $Z$ can be expressed in terms of the mutual information as
\begin{align}
 H(Z) &= H(Z) - H(Z \mid X,N_Z) = I(X, N_Z :Z) = I(X:Z) + I(N_Z:Z \mid X)
\end{align}
For a fixed $\mathbb{P}(Z, X)$, all the SCMs will have the same $H(Z)$ and same $I(X:Z)$, therefore they will all have the same $I(N_Z:Z \mid X)$, i.e. the information that the noise has about $Z$ when $X$ is known. However, we can decompose $I(X,N_Z : Z)$ also into 
\begin{align} \label{eq:characdec} 
 I(X, N_Z :Z) &= I(N_Z:Z) + I(X:Z \mid N_Z).
\end{align}
In \eqref{eq:characdec} only the sum is the same across different SCMs, but each of the two terms differs. This decomposition defines an interpretable characteristic difference of the SCMs, namely whether or not $N_Z$ already provides a significant part of information about $Z$ if {\it $X$ is not known}. 
Here it is important that \eqref{eq:characdec} does not necessarily refer to the response function formulation. Instead, $N_Z$ can be any noise variable that contains \textit{all} other factors in the world that together with $X$ determine $Z$, in addition to potentially parts of the world that have no relevance for $Z$ (since these parts do not affect the terms in \eqref{eq:characdec}). The following result bounds the range in which the decompositions can differ:
\begin{lemma}[range of the two information components]\label{range_it}
For any SCM from $X$ to $Z$ we have 
\begin{align*} 
I(N_Z : Z) & \in [0, H(Z|X) ] \\
I(X:Z \mid N_Z) & \in [ I(X:Z), H(X)]. 
\end{align*} 
\end{lemma}
The proof of Lemma \ref{range_it} can be found in Appendix Sec. \ref{app:proof_lemma_2}.
\begin{example}[independent binaries]\label{ex:binaries} 
For binary $X,Z$, the decomposition \eqref{eq:characdec} 
gets most apparent in the case where $P(Z=1| X=0) = P(Z=1|X=1) = 1/2$ with the two extreme SCMs
where we have either equally weighted mixture of the functions $\mathbf{0},\mathbf{1}$ or an equally weighted mixture of $\mathbf{ID},\mathbf{NOT}$. In the first case, $N_Z$ determines $Z$ entirely, that is, we have 
 \begin{align} 
 I(N_Z: Z) = 1 \quad \hbox{ and } \quad I(X:Z \mid N_Z) = 0.
 \end{align} 
In the second one we obtain 
\begin{align} 
 I(N_Z: Z) = 0 \quad \hbox{ and } \quad I(X:Z \mid N_Z) = 1. 
 \end{align} 
In other words, the two extreme cases of $\lambda_X$ also correspond to the two extreme cases of distributing the information $H(Z)=1$ between the two terms in \eqref{eq:characdec}. 
\end{example}
To interpret $I(X : Z \mid N_Z)$, note that the unconditional mutual information $I(X:Z)$ describes to what extent $X$ controls the statistics of $Z$ in the population mean. The conditional mutual information $I(X : Z \mid N_Z)$ measures the strength of counterfactual influence in the sense of quantifying to what extent variation of $X$ influences $Z$ for every single instance (for which $N_Z$ is fixed).\footnote{Note that we are not confusing correlation and causation here: A priori, strength of dependence does not measure strength of influence, but here this is justified because $X$ and $N_Z$ are the only causes of $Z$. More explicitly, $I(X: Z \mid N_Z)$ measures the strength of the edge $X\to Z$ in the DAG in Fig. \ref{fig:teaser} according to \cite{causalstrength} (see Theorem 4 therein), while in the DAG consisting of $X\to Z$ only, the strength of $X\to Z$ is given by $I(X:Z)$.} For fixed values $n_Z$ of $N_Z$, the conditional mutual information $I(X:Z \mid N_Z = n_Z)$ represents the mutual information between $X$ and $Z$ with respect to a fixed response function. Therefore, it also measures how $Z$ varies for different hypothetical inputs $X$ for each individual unit with $N_Z= n_Z$. We thus state: {\bf Information theoretic measures capture the strength of the influence that $X$ has on $Z$ when adjusting all other factors influencing $Z$.} 

Remarkably, this difference in the decomposition of information is also characteristic for falsifiability of the SCMs $X \rightarrow Z$ using $\mathbb{P}(Y,Z)$. Note that we can assume $Y$ to be a function of $N_Z$ without loss of generality, that is $Y = g(N_Z)$, because we can otherwise define a new noise variable $\tilde{N}_Z:= (N_Z,Y)$ to also include the part of the world that was previously not captured by $N_Z$.
Due to the monotonicity of mutual information 
with respect to functional relations we then have 
\begin{equation} 
I(Y : Z ) \leq I(N_Z : Z). 
\end{equation} 
Therefore, any hypothesised SCM $X \rightarrow Z$ with a corresponding information between $N_Z$ and $Z$ is falsified by $\mathbb{P}(Y,Z)$ when $Y$ contains more information about $Z$ than the hypothesised $N_Z$ is supposed to. Consequently, hypothesised SCMs that assume low values of $I(N_Z : Z)$ are easy to falsify, however in the case that $Y$ contains much less information about $Z$ than the hypothesised $N_Z$ is supposed to, this could be because $Y$ does not tell us enough information about the relevant part of $N_Z$. 

Hence, models with large counterfactual influence in the sense of having large $I(X : Z \mid N_Z)$ and low values of $I(N_Z: Z)$ are easy to falsify. 
This changes, however, when we have access to trivariate observations and use $P(X,Y,Z)$. 
We first observe: 
\begin{lemma}[monotonicity of conditional influence]\label{lem:counterinf} 
For any $Y$ that is a function of $N_Z$ we have 
\[
I(X : Z \mid Y) \leq I(X : Z \mid N_Z). 
\]
\end{lemma} \footnote{See Appendix Section \ref{app:proof_lemma_3} for proof}.
Similar to before, whenever we observe a variable $Y$, for which $I(X : Z \mid Y)$ exceeds the value of $I(X:Z \mid N_Z)$ for the hypothesised SCM, we can falsify this SCM due to Lemma \ref{lem:counterinf}. This is also plausible because the low value of $I(X:Z \mid N_Z)$
corresponds to an insignificant influence of $X$ on $Z$ on the level of individual units, as described above. This is clearly falsified if we discover population subgroups,
defined by different values of $Y$, for which we observe a significant influence of $X$ on $Z$. 

Although the above information theoretic bounds are convenient, they provide in general only {\it necessary} conditions for consistency of $\mathbb{P}(Y,Z)$ with the hypothetical SCM.\footnote{Note that information theoretic bounds based on submodularity of Shannon entropy (which are also not tight) have been derived for deciding whether an observed joint distribution is consistent with a hypothetical latent causal structure \cite{ChavesLMGJS2014}.} However, in the case with binary $X,Z$, the terms $I(N_Z:Z)$ and $I(X: Z \mid N_Z)$ already determine the SCM (with respect
to its response function formulation).

\begin{lemma}[relating $\lambda_X$ with information]\label{lambda_x_it}
For binary $X,Z$, the parameter $\lambda_X$ is linearly related with $I(N_Z:Z)$:
\begin{align}
I(N_Z : Z) = H(Z) - H(X)\left[p_{00} + p_{01} - 2\lambda_X \right].
\end{align}
\end{lemma}\footnote{See Appendix Section \ref{lem:lemma-5} for proof}
 
Hence, large values of $\lambda_X$ correspond to low values of the counterfactual influence $I(X: Z\mid N_Z)$. While information theory provides an elegant framework to falsify SCMs with high dimensional data, it cannot provide tight bounds. To achieve the latter, in the following section we show how to impose counterfactual consistency constraints.

\section{Bounding Probabilities Of Causation}\label{sec:bounding_poc}
Tightening the bounds on the Probabilities of Causation directly translates into a linear programming problem derived from the consistency constraints in Section \ref{sec:causal-marginal}. To see this, note the value of $\lambda_X$ corresponds to the first element of the column vector $\mathbf{A}\mathbf{c}$. This gives us an objective function that is linear in $\mathbf{c}$, and as seen in Section \ref{sec:causal-marginal}, all of the constraints are linear in $\mathbf{c}$ as well. 

As seen in Section \ref{sec:information_theoretic_counterfactuals}, the upper bound on $\lambda_X$ cannot be falsified, so we only solve the minimisation problem to bound PNS. In general, even with two marginals the linear program is challenging to solve symbolically. However, when $\mathbb{P}(Y, Z)$ has degenerate conditionals, e.g. $\mathbb{P}(Z = 0 \mid Y = 0) = 0$ or $\mathbb{P}(Z = 0 \mid Y = 0) = 1$ 
%(with a similar condition on $\mathbb{P}(Z = 0 \mid Y = 1)$)
, we can provide symbolic bounds for PNS.

Such distributions can occur in the context of randomised studies when the disease has a high mortality rate under no treatment, or is the treatment has really high effectiveness. We formally define what it means for $\mathbb{P}(Y, Z)$ to have degenerate conditionals below.

\begin{definition}
A distribution $\mathbb{P}(Y, Z)$ is said to have degenerate conditionals if any one of these conditions hold: $p^{\prime}_{00} = 0$, $p^{\prime}_{01} = 0$, $p^{\prime}_{00} = 1$ or $p^{\prime}_{01} = 1$. Here $p^{\prime}_{00} = \mathbb{P}(Z = 0 \mid Y = 0)$ and $p^{\prime}_{01} = \mathbb{P}(Z = 0 \mid Y = 1)$.
\end{definition}
For settings in which $\mathbb{P}(Y, Z)$ has degenerate conditionals, we provide results on how combining such $\mathbb{P}(Y, Z)$ with $\mathbb{P}(X, Z)$ tightens the bounds on the Probabilities Of Causation. But first, we must check whether $\mathbb{P}(X, Z)$ and $\mathbb{P}(Y, Z)$ are compatible - consistency constraints in Section \ref{sec:causal-marginal} do not provide us with conditions to do so. In the following theorem we show that when $\mathbb{P}(Y, Z)$ has degenerate conditionals, a simple condition can be checked to ensure compatibility.
\begin{theorem}\label{thm:causal_compatibility}
When $\mathbb{P}(Y, Z)$ has degenerate conditionals, conditions derived from imposing $\mathbf{c}$ satisfy $[\mathbf{A}\mathbf{c}]_0 = \lambda^{\max}_X$ in addition to the consistency constraints in Section \ref{sec:causal-marginal} are both necessary and sufficient to ensure $\mathbb{P}(X, Z)$ and $\mathbb{P}(Y, Z)$ are compatible.
\end{theorem}

We now present our tightened bounds on PNS, and consequently, on the Probabilities of Causation in Theorem \ref{thm:pns_bounds} below.

\begin{theorem}\label{thm:pns_bounds}
When $\mathbb{P}(Y, Z)$ has degenerate conditionals, and the conditions outlined in \ref{thm:causal_compatibility} hold, the enforcement of counterfactual consistency tightens the upper bound of PNS as
\begin{align}
\max[0, p_{11} - &p_{10}] \leq PNS \leq \min\{p_{00} - \mathbb{D}_1, p_{11} - \mathbb{D}_0\}
\end{align}
Where $\mathbb{D}_0 \equiv \mathbb{I}(p^{\prime}_{00} = 0)\mathbb{P}(Y = 0) + \mathbb{I}(p^{\prime}_{01} = 0)\mathbb{P}(Y = 1)$ and $\mathbb{D}_1 \equiv \mathbb{I}(p^{\prime}_{00} = 1)\mathbb{P}(Y = 0) + \mathbb{I}(p^{\prime}_{01} = 1)\mathbb{P})(Y = 1)$
\end{theorem}

The new bounds may not be `tight enough', and sometimes we must base decisions on inadequate evidence. In the next section we present approaches to reason in such situations. The proofs of the two Theorems can be found in the Appendix Sec. \ref{prf_thm1} and \ref{prf_thm2} respectively.

\section{Reasoning With Imprecise Probabilities Using MaxEnt}\label{sec:imprecise_probabilities}

There exist many situations in which the bounds on the Probabilities of Causation are non-informative, yet a decision must be made. For example, a clinician running a trial contemplates stopping a trial based on preliminary safety data, and for ethical reasons, doesn't have the time to wait on thorough results to make a decision. To mitigate patient harm, the clinician must decide whether to stop this trial based on insufficient evidence. The theory presented in this section aims to suggest alternative methods to make a calculated decision in such situations of insufficient evidence. To reason under this uncertainty, we suggest adopting a  Maximum Entropy based approach (MaxEnt). We present the details of it, as well as justifications and weaknesses, and leave the choice of the appropriate framework to the domain expert. This approach can be thought of as complimentary to the simplex bounds presented above, i.e. when the bounds aren't informative enough and we are forced to reason nonetheless. 

The set of SCMs over $X$, $Y$ and $Z$ (corresponding to the causal graph in Fig \ref{fig:teaser}) compatible with the observed probabilities forms a convex set, proved in \cite{gresele2022causal}. Each element of this convex set, i.e. $\mathbb{P}(N_Z)$ is a distribution over the $16$ possible functions in the response function formulation. Accordingly, this set contains a unique $\mathbb{P}(N_Z)$ that maximises $H(N_Z)$. We utilise the SCM corresponding to the entropy maximising $\mathbb{P}(N_Z)$, referred to as MaxEnt SCM, in two ways, first to obtain a point estimate of $\lambda_X$, and second, to quantify the strength of evidence of various datasets $\mathbb{P}(Z, Y)$. 

\textbf{Point estimate of $\lambda_X$:} The SCM maximising the entropy amounts to a single value of $\lambda_X$, which can be seen as a `good guess', in the same way as in the general context of MaxEnt, the distribution maximizing the entropy subject to the given constraint, can be seen as a `good guess', if one is forced to assume a unique distribution. For further discussions on the justification of MaxEnt we refer to \cite{jaynes2003probability, shore1980axiomatic,uffink1995can}, 
and also to \cite{grunwald2004game}, where MaxEnt follows from the minimisation of a worst case loss
in a game.\footnote{Since MaxEnt is equivalent to minimizing KL distance to the uniform distribution, one can easily generalise it to arbitrary priors over functions.
 For SCMs with many inputs we may want to give preference to `simple' functions, which conflicts with the implicit uniform prior over all
functions MaxEnt relies on. 
}
  
\textbf{Strength of evidence:} When the entropy of the MaxEnt SCM is small, the space of admissible SCMs is small since the MaxEnt distribution of a large volume is large (note the converse doesn't hold\footnote{A large value of the MaxEnt distribution does {\it not} tell us that the space of admissible solutions is large. Instead, it could contain a single distribution of high entropy}). Consequently, observing $P(Y,Z)$ that significantly reduces the entropy of the MaxEnt SCM from $X,Y$ to $Z$ compared to when nothing is known about $P(Y,Z)$, tells us that the additional observation has significantly reduced the space of possible SCMs. 
In this sense, significant reduction of entropy also implies that a large space of SCMs that were possible prior to observing $P(Y,Z)$ would now be falsified. 

This property of MaxEnt motivates us to use it as a metric to quantify the strength of evidence provided by various datasets. Our solution is to rank the various available datasets $\mathbb{P}(Z, Y)$ based on the magnitude of entropy reduction. Here, `strong evidence' means significant reduction of uncertainty (entropy) about the response function, and as such, falsification of a large space of SCMs. Of course, datasets where $Y$ has little information about $Z$ do not provide us with much information about the relation between $X$ and $Z$. Next, we demonstrate the properties of MaxEnt through numerical simulations, using the python package \textit{scipy.optimize.minimize}\footnote{\url{https://docs.scipy.org/doc/scipy/reference/generated/scipy.optimize.minimize.html}} to solve the dual version \footnote{Solving the dual of the MaxEnt optimization problem reduces the parameters to optimise from 15 to 4 (the number of constraints)} of this optimization problem, a derivation of which can be found in \cite{li2004duality}. 

\begin{figure}[H]
\centering
\includegraphics[scale=0.5]{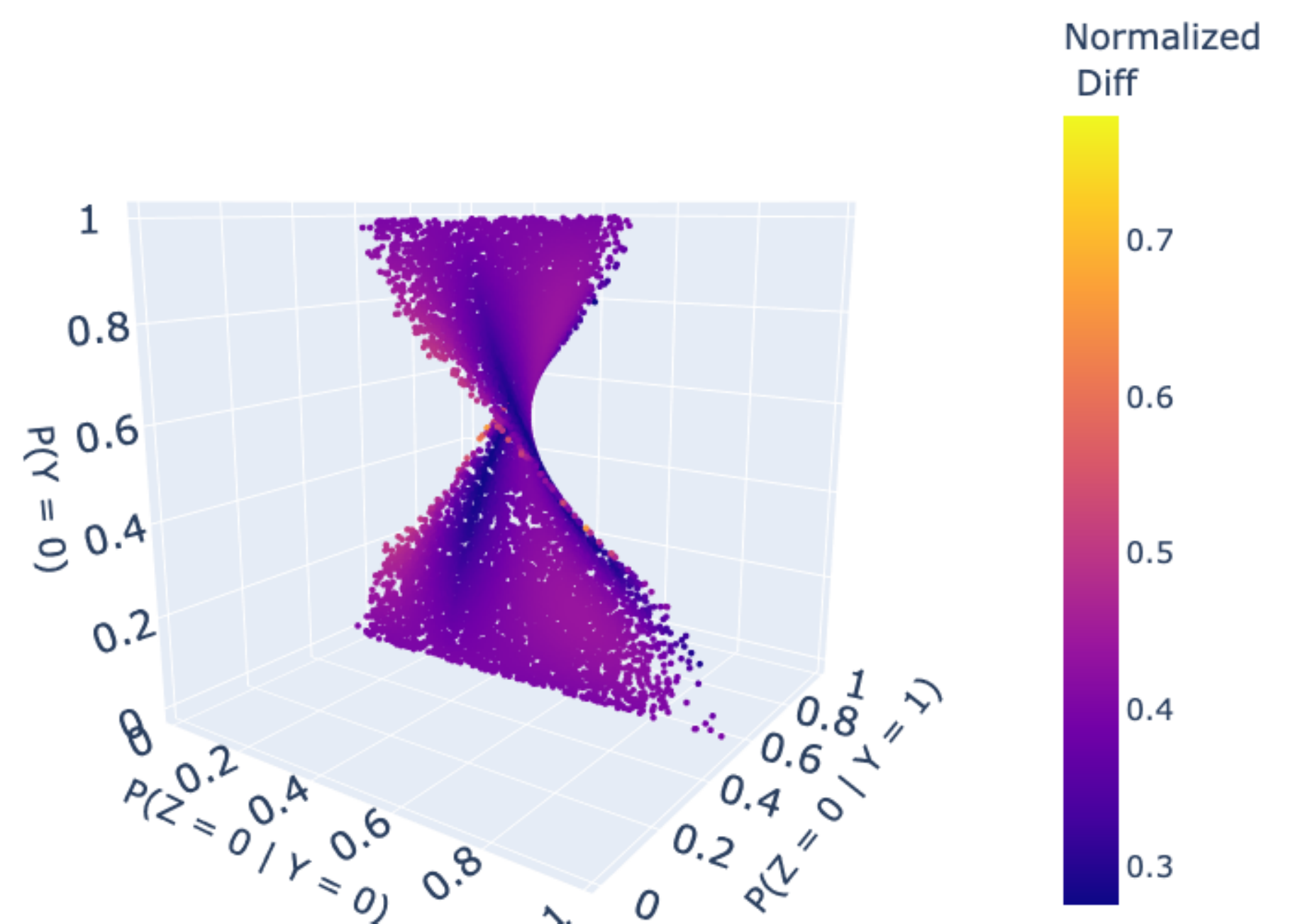}
\caption{\textbf{MaxEnt estimates of $\lambda_X$}: the difference of the value $\lambda_X$ estimated by MaxEnt, and the $\lambda_X^{\text{max}}$, normalised by the range of possible $\lambda$ s. Here we observe that the $\lambda_X$ estimated by MaxEnt is most of the times not far from the middle of the interval $[\lambda_X^{\text{min}}, \lambda_X^{\text{max}}]$, which corresponds to the normalised difference $0.5$.}\label{maxent_lambdax}
\end{figure}

\begin{figure}[H]
\centering
\subfigure[] {\includegraphics[scale=0.25]{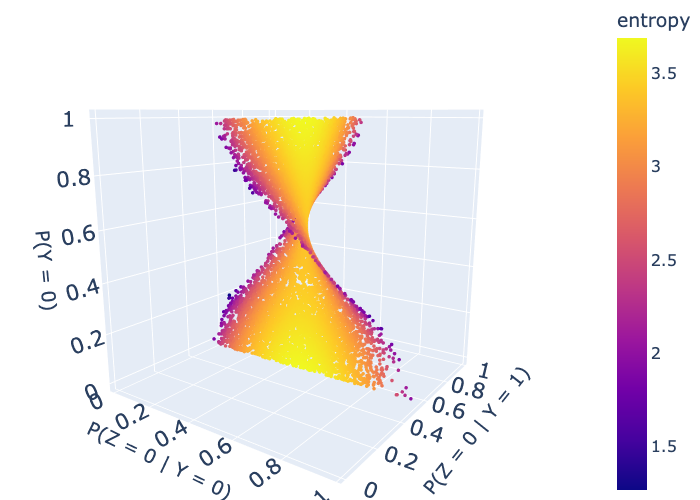} \label{maxent_strength}}\quad \subfigure[]{\centering\includegraphics[scale=0.25]{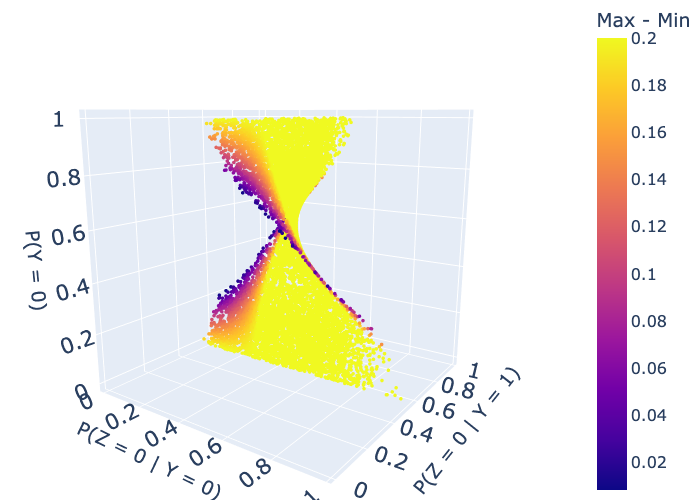}\label{lambda_range}}\quad \subfigure[]{\centering\includegraphics[scale=0.3]{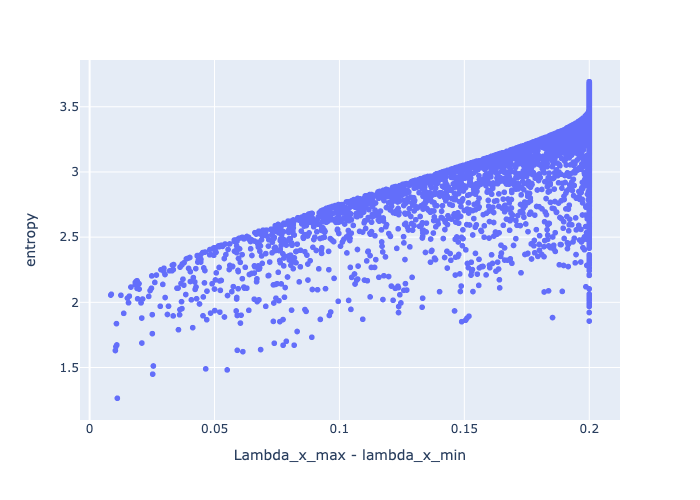}\label{entropy_vs_lambdaxs}}
\caption{For fixed values of $p_{00}$, $p_{01}$ and $\mathbb{P}(X = 0)$, we vary $p^{\prime}_{00}$, $p^{\prime}_{01}$ and $\mathbb{P}(Y = 0)$ and we plot (a) \small \textbf{MaxEnt as strength of evidence}: the entropy of the maximum entropy SCM for the defined family of SCMs. Values of $p^{\prime}_{00}$, $p^{\prime}_{01}$ and $\mathbb{P}(Y = 0)$ that decrease the entropy the most are considered the strongest forms of evidence. (b) the \small \textbf{Range of $[\lambda_X^{\text{min}}, \lambda_X^{\text{max}}]$}, and (c) Entropy of the joint SCM vs the interval $[\lambda_X^{\text{min}}, \lambda_X^{\text{max}}]$.}
\end{figure}

All of the figures that follow are generated for a fixed $\mathbb{P}(X, Z)$ and we vary $\mathbb{P}(Y, Z)$. In figures \ref{maxent_strength} and \ref{lambda_range} we see that low entropy of the MaxEnt SCM tends to correspond to regions with low values of the range of $\lambda_X$. More precisely, in \ref{entropy_vs_lambdaxs} we see that in general this not a monotonic relationship - because, as expected, not all $\mathbb{P}(Z, Y)$ that restrict the space of SCMs are capable of restricting the range of $\lambda_X$. Note that $\lambda^{\max}_X - \lambda^{\min}_X = 0.2$ in Figure \ref{entropy_vs_lambdaxs} is the size of the interval $[\lambda^{\max}_X, \lambda^{\min}_X]$ without restrictions from $\mathbb{P}(Y, Z)$. For point estimates of $\lambda_X$, Figure \ref{maxent_lambdax} shows that the value of $\lambda_X$ estimated by MaxEnt is in the interior of $[\lambda^{\max}_X, \lambda^{\min}_X]$, which is not the middle of interval, but here not too far from it. 

%\begin{SCfigure}[0.99][h]
%\caption{\small \textbf{MaxEnt estimates of $\lambda_X$}: the difference of the value $\lambda_X$ estimated by MaxEnt, and the $\lambda_X^{\text{max}}$, normalised by the range of possible $\lambda$ s. Here we observe that the $\lambda_X$ estimated by MaxEnt is most of the times not far from the middle of the interval $[\lambda_X^{\text{min}}, \lambda_X^{\text{max}}]$, which corresponds to the normalised difference $0.5$.}
%\includegraphics[scale=0.25]{plots/lambda_x_vs_pzy.pdf}\label{maxent_lambdax}
%\end{SCfigure}

% \subsection{A Bayesian Inference Framework}
%\subfile{sections/bayesian_confirmation_theory}

\section{Discussion}\label{sec:discussion}
Having presented various approaches for counterfactual inference, we briefly highlight some key discussion points.

\paragraph{Connection to data fusion} A similar line of work on \emph{data fusion} \cite{bareinboim2016causal, hunermund2019causal} aims to identify interventional effects by combining multiple experimental and observational datasets. We draw a few key distinctions between this line of work and ours. First, this approach assumes access to distributions is over the \emph{same} set of random variables, with \emph{different} variables being intervened on. In contrast, we assume access only to distributions over \emph{different} sets of random variables, i.e. \emph{same} outcome but different \emph{treatments}. Second, since we assume a particular causal graph given in Fig. \ref{fig:teaser} over binary random variables $X$, $Y$ and $Z$, we are able to provide tighter bounds on the Probabilities of Causation than the aforementioned work. 

%, and describe falsification of Probabilities of Causation 
%and SCMs from an information theoretic perspective that generalizes beyond our simple scenario and renders falsification of SCMs via additional variables more interpretable.   
%Next, our novel information theoretic approach allows us to reason about falsifying SCMs using high dimensional datasets, which are become more common in the era of big data - e.g. EHR\cite{johnson2020mimic} and genomics datasets\cite{hutter2018cancer}. \numair{This is a little buzzwordy, but nevertheless true statement}

\paragraph{Binary variables} While we do restrict our attention to binary variables, this case captures an important real world scenario nevertheless. For example, the disease Rheumatoid Arthritis, is treated using various multiple pharmaceutical drugs - two such drugs that have independent trials conducted to their study Baricitinib \cite{taylor2017baricitinib} and Filgotinib \cite{genovese2019effect} . Both these trials measure different treatments but the same outcome (ACR20 response criteria for Rheumatoid Arthritis \cite{felson1995american}). This is a characteristic case where our method enables clinicians to leverage all of these independent trials to strengthen counterfactual inference and make decisions about the effectiveness of the drugs. 

\paragraph{Bounds with degenerate conditionals} It may seem that asking for marginals with `degenerate conditionals' (being at the boundary of the set of distributions, on which parts of our results rely on) is a strong requirement. However, there are many trials whose observed data distribution approximates a distribution with degenerate conditionals. For example, the BioNtech trial for the BNT162b2 vaccine \cite{polack2020safety} with $95\%$ efficacy could be used as an approximation to utilise our tightened bounds.

%In this trail, $T = 0$ corresponds to placebo and $T = 1$ corresponds to the  candidate, and the outcome $Z = 1$ represents laboratory confirmed Covid-19. 
%\numair{$\mathbb{P}(Z = 0 \mid T = 1) = 0.95$, where $T$ represents whether the vaccine was administered or not, and $Z$ represents whether the patient contracted COVID or not}. 

\paragraph{Future work on multiple marginal datasets}  
As mentioned before in Section \ref{sec:information_theoretic_counterfactuals}, the information theoretic perspective is also of use in scenarios with multiple datasets containing different subsets of high dimensional random variables $\{X, Z, Y_1, . . . , Y_d\}$. 
%and multiple datasets containing different subsets of $\{X, Z, Y_1, . . . , Y_d\}$. 
Following ideas in \cite{ChavesLMGJS2014} based on submodularity of Shannon entropy, the
entropy and mutual information of the subsets entail necessary conditions for the existence of a joint distribution of $\{N_Z,X,Z,Y_1,...,Y_d\}$ that is consistent with the hypothetical counterfactual influence $I(X : Z \mid N_Z )$. This way, multiple additional variables together can potentially falsify the SCM from $X$ to $Z$ - we leave this to future work.

\paragraph{Conclusion}
In this paper we presented several approaches to get evidence on the Probabilities of Causation via merging datasets with different treatment variables and the same outcome variable. To this end, we tightened existing bounds on the Probabilities of Causation, and developed a new information theoretic perspective that confers interpretability and tractability advantages for falsification of SCMs. Additionally, in the absence of additional evidence, we provide a Maximum Entropy based framework for counterfactual inference. Our approach generalises to multiple variables and counterfactual statements (rung 3 in Pearl’s ladder of causation) other than Probabilities of Causation, remarks about which can be found in Section \ref{additional_counterfactuals} of the Appendix. 

%In this paper we presented several approaches to get evidence on the Probabilities of Causation via merging datasets with different treatment variables and the same outcome variable. To this end, we showed how to use the Causal Marginal Problem to tighten existing bounds on the Probabilities of Causation, and developed a new information theoretic perspective with advantages regarding insights and tractability in high dimensions for falsification of SCMs. Additionally, in the absence of additional evidence, we provide a Maximum Entropy based as well as a Bayesian framework for counterfactual inference. Our approach generalises to multiple variables and counterfactual statements (rung 3 in Pearl’s ladder of causation) other than Probabilities of Causation, remarks about which can be found in Section xx of the Appendix. 

\newpage
{\Huge \textbf{Appendix}}
\section{Parameterisation of SCM $X \rightarrow Z$ from $\mathbb{P}(Z, X)$}\label{lambda_x_derivation}
The SCM from $X \rightarrow Z$ when $X$ and $Z$ are both binary random variables is represented using the response function parameterisation as
\begin{align}
    X := N_X\\
    Z := f_{N_Z}(X)
\end{align}
Under the response function parametrisation, every SCM is uniquely characterised by the distribution of its noise variables, $\mathbb{P}(N_Z, N_X)$. Here $N_Z$ is a random variable that takes a function value from the set $\left\{\mathbf{0}, \mathbf{1}, \mathbf{ID}, \mathbf{NOT} \right\}$, and $N_X$ is an binary random variable that directly generates the value of $X$. While $\mathbb{P}(N_X)$ is identified from $\mathbb{P}(Z, X)$ as $\mathbb{P}(X)$, $N_Z$ is not. However we can still obtain bounds on $\mathbb{P}(N_Z)$ using $\mathbb{P}(Z, X)$ as follows. 

Let $\mathbb{P}(N_Z)$ be parameterised as $[a_0, a_1, a_2, a_3]$ where each $a_i$ represents the probability mass placed on the respective functions in $\left\{\mathbf{0}, \mathbf{1}, \mathbf{ID}, \mathbf{NOT} \right\}$. Since the probability of observing $Z = 0$ when $X = 0$ corresponds to the probability mass placed on the $\mathbf{0}$ and $\mathbf{NOT}$ function, $\mathbb{P}(Z, X)$ can be used to constrain $\mathbb{P}(N_Z)$ as follows
\begin{align}
 	a_0 + a_2 &= \mathbb{P}(Z = 0 \mid X = 0)\\
	a_0 + a_3 &= \mathbb{P}(Z = 0 \mid X = 1)\\
	a_0 + a_1 + a_2 + a_3 &=  1
\end{align}

With three equations constraining four free variables, this system of equations has one free parameter. We choose $a_0$ as the free parameter and denote it by $\lambda_X$. This allows us to re-express the above equations in terms of $\lambda_X$ as 
\begin{align}
    a_2 &= \mathbb{P}(Z = 0 \mid X = 0) - \lambda_X\\
    a_3 &= \mathbb{P}(Z = 0 \mid X = 1) - \lambda_X\\
    a_1 &= 1 - \mathbb{P}(Z = 0 \mid X = 0) - \mathbb{P}(Z = 0 \mid X = 1) + \lambda_X
\end{align}

Since each of the $a_i$'s must be valid probabilities i.e. be non-negative as well less than or equaL to $1$, we are able to obtain the range for $\lambda_X$ as 
\begin{align}
\max\{0, p_{00} + p_{01} - 1\} \leq \lambda_X \leq \min\{p_{00}, p_{01}\}
\end{align}
Where $p_{00} = \mathbb{P}(Z = 0 \mid X = 0)$ and $\mathbb{P}(Z = 0 \mid X = 1)$. The final parameterisation for any SCM consistent with $\mathbb{P}(Z, X)$ is given as
\begin{align}
    \mathbf{a}(\lambda_X) = \begin{pmatrix}0 \\ 1 - p_{00} - p_{01}\\ p_{00} \\ p_{01}\end{pmatrix}+ \lambda_X\begin{pmatrix}1 \\ 1 \\ -1 \\ -1\end{pmatrix}
\end{align}

\section{Justification For $\lambda_X$ Capturing Counterfactual Influence}\label{lambda_x_counterfactual}

The counterfactual sentence "Z would have been different if X had been different, given that we observe a particular value of Z and X" can be presented as the probability $\mathbb{P}(Z_{x^{\prime}} = z^{\prime} \mid Z = z, X = x)$. Since the $\mathbf{0}$ and $\mathbf{1}$ functions are unable of generating differing outputs with differing inputs, this counterfactual probability must be proportional to the weight placed on the $\mathbf{ID}$ and $\mathbf{NOT}$ functions, which are the only functions capable of changing the outputted value $Z$ upon a change in $X$. Hence $\mathbb{P}(Z_{x^{\prime}} = z^{\prime} \mid Z = z, X = x) \propto  p_{00} + p_{01} - 2\lambda_X$. Both $p_{00}$ and $p_{01}$ are fixed functions of $\mathbb{P}(Z, X)$, so the counterfactual influence of SCMs consistent with $\mathbb{P}(Z, X)$ only varies with $\lambda_X$. 

\section{Proof Of Lemma \ref{lem:link_lambda}}
PNS is expressed as the joint probability of two events, $Z_x = z$ and $Z_{x^{\prime}}= z^{\prime}$, i.e. $\mathbb{P}(Z_{x} = z, Z_{x^{\prime}} = z^{\prime})$ where $z = 1$, $z^{\prime} = 0$ and the same holds for $x$ and $x^{\prime}$. Therefore, PNS corresponds to the probability mass placed on the $\mathbf{ID}$ function, since this is the only function capable of outputting $Z = 1$ when $X = 1$ and $Z = 0$ when $X = 0$. Under the parameterisation of $\mathcal{M}_X$ presented in Section \ref{lambda_x_derivation}, PNS will equal $p_{00} - \lambda_X$. 

Now the bounds on PNS are recovered by substituting in the bounds on $\lambda_X$ presented in Section \ref{lambda_x_derivation}. The lower bound on $\lambda_X$ will translate into the upper bound on PNS and vice versa. This yields the bounds
\begin{align}
    \max[0, p_{11} - p_{10}] \leq PNS \leq \min[p_{00}, p_{11}]
\end{align}
Which recovers the bounds from \cite{tian2000probabilities} on the PNS. 

\section{Proof of Lemma \ref{range_it}}\label{app:proof_lemma_2}

To prove this lemma, we first establish and proof the following lemma.
\begin{lemma}\label{lem:indcond} 
 For any three random variables $U,V,W$ with $U \independent V$ we have 
 \begin{align}
 I(U: W) \leq I(U : W \mid V).  
 \end{align}
 \end{lemma} 
\textbf{Proof}: Using properties of conditional mutual information, $I(U : W\mid V)$ can be written as
\begin{align}
I(U : W \mid V) &= H(U \mid V) - H(U \mid V, W)\\
\intertext{Since $U \independent V$, we can rewrite the above as}
I(U : W \mid V) &= H(U) - H(U \mid V, W)\\
\end{align}
Similarly, one can rewrite $I(U: W)$ is written as
\begin{align}
I(U: W) &= H(U) - H(U \mid W)
\end{align}
Since $H(U \mid V, W) \leq H(U \mid W)$, the Lemma holds.\\
With the presentation of the above Lemma, we are now ready to proceed with the proof.\\

\textbf{Proof for Lemma \textcolor{red}{3} \textcolor{green}{2}}: We first show the bounds for $I(X:Z \mid N_Z)$. The lower bound on $I(X:Z \mid N_Z)$ follows from the independence of $X$ and $N_Z$ combined with Lemma \ref{lem:indcond}. The upper bound on $I(X:Z \mid N_Z)$ can be seen as follows.
\begin{align}
I(X:Z \mid N_Z) = H(X|N_Z) - H(X| N_Z,Z) \leq H(X|N_Z) \leq H(X) 
\end{align}

Therefore, the bounds on $I(X:Z \mid N_Z)$ are 
\begin{align}
I(X: Z) \leq I(X:Z \mid N_Z) \leq H(X)
\end{align}

To demonstrate the bounds on $I(N_Z : Z)$, we note the lower bound is obtained from the non-negative property of mutual information, and is attained when $N_Z \independent Z$. For the upper bound, note
\begin{align}
I(N_Z : Z) = H(Z) - I(X:Z \mid N_Z) &\leq    H(Z) - I(X:Z) = H(Z|X)\\
I(N_Z : Z) &\leq H(Z\mid X)\\
0 \leq I(N_Z:Z) &\leq H(Z\mid X)
\end{align}

\section{Derivation For Example \ref{ex:binaries}}

To derive the results presented, we first note that $\mathbb{P}(Z = 0 \mid X = 0) = \mathbb{P}(Z = 0 \mid X = 1) = \frac{1}{2}$ implies $\mathbb{P}(Z = 0) = \frac{1}{2}$, and consequently $H(Z) = 1$. Next, $\mathbb{P}(Z = 0 \mid X = 0) = \mathbb{P}(Z = 0 \mid X = 1) = \frac{1}{2}$, also sets the range of $\lambda_X$ as $\lambda_X \in [0, \frac{1}{2}]$. The upper bound $\lambda_X = \frac{1}{2}$ corresponds to a model with $0$ counterfactual influence, and this model places equal weights on the $\mathbf{0}$ and $\mathbf{1}$ function. So, $\mathbb{P}(N_Z)$ can be written as

\begin{align}
\mathbb{P}(N_Z = 0) = \frac{1}{2}\\
\mathbb{P}(N_Z = 1) = \frac{1}{2}\\
\mathbb{P}(N_Z = 2) = 0\\
\mathbb{P}(N_Z = 3) = 0
\end{align} 
Given the distribution of $\mathbb{P}(N_Z)$ above, we can calculate the distribution of $\mathbb{P}(Z, N_Z)$ as
\begin{align}
\mathbb{P}(Z = 0, N_Z = 0) = \frac{1}{2}\\
\mathbb{P}(Z = 0, N_Z = 1) = 0\\
\mathbb{P}(Z = 0, N_Z = 2) = 0\\
\mathbb{P}(Z = 0, N_Z = 3) = 0\\
\mathbb{P}(Z = 1, N_Z = 0) = 0\\
\mathbb{P}(Z = 1, N_Z = 1) = \frac{1}{2}\\
\mathbb{P}(Z = 1, N_Z = 2) = 0\\
\mathbb{P}(Z = 1, N_Z = 3) = 0
\end{align}
Using the calculated dsitrbution the formula for $I(N_Z:Z)$ gives us $I(N_Z:Z) = 1$. And utilising the decomposition with $H(Z)=1$ proves $I(X :Z \mid N_Z) = 0$ for the SCM with $\lambda_X = \frac{1}{2}$.\\
For the SCM with $\lambda_X = 0$, $\mathbb{P}(N_Z)$ will place equal weight on the $\mathbf{ID}$ and $\mathbf{NOT}$ function. As a result, $\mathbb{P}(Z, N_Z)$ will equal
\begin{align}
\mathbb{P}(Z = 0, N_Z = 0) = 0\\
\mathbb{P}(Z = 0, N_Z = 1) = 0\\
\mathbb{P}(Z = 0, N_Z = 2) = \frac{1}{2}\mathbb{P}(X = 0) = \frac{1}{4}\\
\mathbb{P}(Z = 0, N_Z = 3) = \frac{1}{2}\mathbb{P}(X = 1)  = \frac{1}{4}\\
\mathbb{P}(Z = 1, N_Z = 0) = 0\\
\mathbb{P}(Z = 1, N_Z = 1) = 0\\
\mathbb{P}(Z = 1, N_Z = 2) = \frac{1}{2}\mathbb{P}(X = 1) = \frac{1}{4}\\
\mathbb{P}(Z = 1, N_Z = 3) = \frac{1}{2}\mathbb{P}(X = 0) = \frac{1}{4}
\end{align}

So, $I(N_Z:Z) = 0$, and from the decomposition, $I(X :Z \mid N_Z) = 1$.

\section{Proof Of Lemma \ref{lem:counterinf}}\label{app:proof_lemma_3}
Since $X \independent N_Z$ and $Y = g(N_Z)$, we also have $X\independent (Y,N_Z)$. An application of the weak union graphoid axiom gives $X\independent N_Z \mid Y$.
Applying Lemma \ref{lem:indcond} to the  variables $X,N_Z,Z$ with the conditional distribution given $Y$, yields
$I(X:Z \mid Y ) \leq I(X: Z \mid Y,N_Z) = I(X: Z \mid N_Z)$. 

\section{Proof Of Lemma \ref{lambda_x_it}}\label{lem:lemma-5}
By definition of mutual information, we have  $I(N : Z) =  -H(Z \mid N_Z) + H(Z)$. 
%we have
%\begin{align}
%    I(N:Z) &= \sum_{Z}\sum_{N_Z} \mathbb{P}(Z, N_Z)\log{\frac{\mathbb{P}(N_Z, Z)}{\mathbb{P}(N_Z)\mathbb{P}(Z)}}
%    \intertext{Which can be further simplified as}
%    &= \sum_{Z}\sum_{N_Z} \mathbb{P}(Z, N_Z)\log{\frac{\mathbb{P}(Z \mid N_Z)\mathbb{P}(N_Z)}{\mathbb{P}(N_Z)\mathbb{P}(Z)}}\\
%    &= \sum_{Z}\sum_{N_Z} \mathbb{P}(Z, N_Z)\log{\frac{\mathbb{P}(Z \mid N_Z)}{\mathbb{P}(Z)}}\\
%    &= \sum_{Z}\sum_{N_Z} \mathbb{P}(Z, N_Z)\log{\mathbb{P}(Z \mid N_Z)} - \sum_{Z}\sum_{N_Z} \mathbb{P}(Z, N_Z)\log{\mathbb{P}(Z)}\\
%    &= \sum_{Z}\sum_{N_Z} \mathbb{P}(Z, N_Z)\log{\mathbb{P}(Z \mid N_Z)} - \sum_{Z} \mathbb{P}(Z)\log{\mathbb{P}(Z)}\\
%    &= \sum_{Z}\sum_{N_Z} \mathbb{P}(Z, N_Z)\log{\mathbb{P}(Z \mid N_Z)} + H(Z)\\
%    &=
%\end{align}

Given $\mathbb{P}(Z, X)$, we know $\mathbb{P}(Z)$ through marginalisation, consequently it is known and fixed for every SCM. We are left to calculate $H(Z \mid N_Z)$, for which we need to calculate $\mathbb{P}(Z \mid N_Z)$ and $\mathbb{P}(Z, N_Z)$, and are enumerated below
\begin{align}
\intertext{For $N_Z = 0$}
    \mathbb{P}(Z = 0 \mid N_Z = 0) &= 1\\
    \mathbb{P}(Z = 1 \mid N_Z = 0) &= 0\\
    \mathbb{P}(Z = 0, N_Z = 0) &= \lambda_X\\
    \mathbb{P}(Z = 1, N_Z = 0) &= 0\\
    \intertext{For $N_Z = 1$}
    \mathbb{P}(Z = 0 \mid N_Z = 1) &= 0\\
    \mathbb{P}(Z = 1 \mid N_Z = 1) &= 1\\
    \mathbb{P}(Z = 0 , N_Z = 1) &= 0\\
    \mathbb{P}(Z = 1, N_Z = 1) &= 1 - p_{00} - p_{01} + \lambda_X\\
    \intertext{For $N_Z = 2$}
    \mathbb{P}(Z = 0 \mid N_Z = 2) &= \mathbb{P}(X = 0)\\
    \mathbb{P}(Z = 1 \mid N_Z = 2) &= \mathbb{P}(X = 1)\\
    \mathbb{P}(Z = 0, N_Z = 2) &= \mathbb{P}(X = 0)\times(p_{00} - \lambda_X)\\
    \mathbb{P}(Z = 1, N_Z = 2) &= \mathbb{P}(X = 1)\times (p_{00} - \lambda_X)
    \intertext{For $N_Z = 3$}
    \mathbb{P}(Z = 0 \mid N_Z = 3) &= \mathbb{P}(X = 1)\\
    \mathbb{P}(Z = 1 \mid N_Z = 3) &= \mathbb{P}(X = 0)\\
    \mathbb{P}(Z = 0, N_Z = 3) &= \mathbb{P}(X = 1)\times (p_{01} - \lambda_X)\\
    \mathbb{P}(Z = 1, N_Z = 3) &= \mathbb{P}(X = 0)\times (p_{01} - \lambda_X)\\
\end{align}
Using this to compute $H(Z\mid N_Z)$, note that the terms with $N_Z = 0$ and $N_Z =1$ will cancel in the expression for $H(Z \mid N_Z)$, leaving us with 
\begin{align}
H(Z \mid N_Z) &= -\mathbb{P}(X = 0)\times(p_{00} - \lambda_X)\log{\mathbb{P}(X = 0)} -\mathbb{P}(X = 1)\times (p_{00} - \lambda_X)\log{\mathbb{P}(X = 1)}\\
&-\mathbb{P}(X = 1)\times (p_{01} - \lambda_X)\log{\mathbb{P}(X = 1)} - \mathbb{P}(X = 0)\times (p_{01} - \lambda_X)\log{\mathbb{P}(X = 0)}\\
&= H(X)[p_{00} - p_{01} - 2\lambda_X]
\end{align}
Substituting this into the expression of $I(N_Z : Z)$
\begin{align}
I(N_Z:Z)= -H(X)\left[p_{00} + p_{01} - 2\lambda_X \right] + H(Z)
\end{align}

\section{Details On Convex Polytope $\mathcal{C}$}\label{convex-polytope}

Here we present the equality and inequality constraints that describe $\mathcal{C}$. $\mathcal{C}$ is the set of values of $\mathbf{c}$ that satisfy the following constraints, and a derivation for this can be found in \cite{gresele2022causal}.
\begin{align}
    \Tilde{\mathbf{A}} \mathbf{c} &\preceq \Tilde{\mathbf{b}}\\
    \Tilde{\mathbf{C}} \mathbf{c} &= \Tilde{\mathbf{d}} 
\end{align}
We define the matrices $\Tilde{\mathbf{A}}$, $\Tilde{\mathbf{b}}$, $\Tilde{\mathbf{C}}$ and $\Tilde{\mathbf{d}}$ below. The notation $\mathbf{a} \preceq \mathbf{b}$, denotes $a_i \leq b_i$ for all entries.

\paragraph{Inequality constraints.}
Here we use $\mathbf{A}$ and $\mathbf{B}$ as defined in \cite{gresele2022causal} Equation 15., and $\Tilde{\mathbf{A}}$ and $\Tilde{\mathbf{b}}$ are given as
\begin{align}
    \Tilde{\mathbf{A}} := \begin{pmatrix} \mathbf{A} \\	- \mathbf{A} \\ \mathbf{B} \\ - \mathbf{B} \\ -\mathbb{I} \end{pmatrix}, \qquad    \Tilde{\mathbf{b}} := \begin{pmatrix} \Tilde{\mathbf{a}}_1 \\ \Tilde{\mathbf{a}}_2 \\ \Tilde{\mathbf{b}}_1 \\ \Tilde{\mathbf{b}}_2 \\ \mathbf{0} \end{pmatrix}
\end{align}

To define $\Tilde{\mathbf{a}}_1$, $\Tilde{\mathbf{a}}_2$, $\Tilde{\mathbf{b}}_1$, $\Tilde{\mathbf{b}}_2$ we first define $\mathbf{a}_0$ and $\mathbf{b}_0$ below
\begin{align}
\mathbf{a}_0 = \begin{pmatrix} 0\\  1-P(Z=0|X=0)-P(Z=0|X=1)\\ P(Z=0|X=0)\\ P(Z=0|X=1) \end{pmatrix}\\
\mathbf{b}_0 = \begin{pmatrix} 0\\  1-P(Z=0|Y=0)-P(Z=0|Y=1)\\ P(Z=0|Y=0)\\ P(Z=0|Y=1) \end{pmatrix}\\
\end{align}

Now, $\Tilde{\mathbf{a}}_1$, $\Tilde{\mathbf{a}}_2$, $\Tilde{\mathbf{b}}_1$, $\Tilde{\mathbf{b}}_2$ are defined as
\begin{align}
\Tilde{\mathbf{a}}_1 =  \mathbf{a}_0 + \begin{pmatrix} \lambda^{\max}_X\\ \lambda^{\max}_X \\ -\lambda^{\min}_X\\ -\lambda^{\min}_X \end{pmatrix} 
\quad \Tilde{\mathbf{a}}_2 =  -\mathbf{a}_0 + \begin{pmatrix} -\lambda^{\min}_X\\ -\lambda^{\min}_X \\ \lambda^{\max}_X\\ \lambda^{\max}_X \end{pmatrix} \\
\Tilde{\mathbf{b}}_1 =  \mathbf{b}_0 + \begin{pmatrix} \lambda^{\max}_Y\\ \lambda^{\max}_Y \\ -\lambda^{\min}_Y\\ -\lambda^{\min}_Y \end{pmatrix} 
\quad \Tilde{\mathbf{b}}_2 =  -\mathbf{b}_0 + \begin{pmatrix} -\lambda^{\min}_Y\\ -\lambda^{\min}_Y \\ \lambda^{\max}_Y\\ \lambda^{\max}_Y \end{pmatrix} 
\end{align}

And $\mathbf{I}$ is a $16$-dimensional vector containing all $1$'s and $\mathbf{0}$ is a 16 dimensional vector of $0$'s.
\paragraph{Equality Constraints}
To define the equality constraints, we define $\Tilde{\mathbf{C}}$ and $\Tilde{\mathbf{d}}$ as 
\begin{align}
    \Tilde{\mathbf{C}} := \begin{pmatrix} \Tilde{\mathbf{C}}_{\mathbf{A}}\\
        \Tilde{\mathbf{C}}_{\mathbf{B}} \\
        \Tilde{\mathbf{C}}_{\mathbf{1}}
    \end{pmatrix} \qquad
    \Tilde{\mathbf{d}} := \begin{pmatrix}        \Tilde{\mathbf{d}}_{\mathbf{a}_0}\\        \Tilde{\mathbf{d}}_{\mathbf{b}_0}\\        1    \end{pmatrix}
\end{align}

Where
\begin{align}
\Tilde{\mathbf{C}}_\mathbf{A}:= \begin{pmatrix}  1 & -1 & 0 & 0 \\  1 & 0 & 1 & 0\\ 1 & 0 & 0 & 1 \end{pmatrix}\mathbf{A}, \quad
\Tilde{\mathbf{d}}_{\mathbf{a}_0} :=\begin{pmatrix}   1 & -1 & 0 & 0 \\ 1 & 0 & 1 & 0\\ 1 & 0 & 0 & 1 \end{pmatrix}\mathbf{a}0\\
\Tilde{\mathbf{C}}_\mathbf{B}:= \begin{pmatrix} 1 & -1 & 0 & 0 \\ 1 & 0 & 1 & 0\\ 1 & 0 & 0 & 1\end{pmatrix}\mathbf{B} \quad 
\Tilde{\mathbf{d}}_{\mathbf{b}_0} :=\begin{pmatrix} 1 & -1 & 0 & 0 \\ 1 & 0 & 1 & 0\\ 1 & 0 & 0 & 1\end{pmatrix}\mathbf{b}_0
\end{align}
And $\Tilde{\mathbf{C}}_{\mathbf{1}}$ is a 16-dimensional row vector of $1$'s.

\section{Proof For Theorem \ref{thm:causal_compatibility}}\label{prf_thm1}

In addition to this re-parameterisation, this proof utilises a Lemmas and Proposition that we present and prove first, and then use them to prove the statement of the Theorem \ref{thm:causal_compatibility}. We begin with a Lemma involving the connection between distributions with degenerate conditionals and special families of SCMs. 
\begin{lemma}\label{degen_cases}
Distributions $\mathbb{P}(Y, Z)$ with degenerate conditionals correspond to families of SCMs where $\lambda^{\min}_Y = \lambda^{\max}_Y$.
\end{lemma}
\paragraph{Proof} This is proved on a case by case basis as follows, where $p^{\prime}_{00}$ denotes $\mathbb{P}(Z = 0 \mid Y = 0)$ and $p^{\prime}_{01} = \mathbb{P}(Z = 0 \mid Y = 1)$:
\begin{itemize}
    \item When $p^{\prime}_{00} = 0$, then $\lambda^{\max}_Y = 0$. $p^{\prime}_{00} + p^{\prime}_{01} - 1$ will at most equal $0$. Therefore $\lambda^{\min}_Y = 0$. So $\lambda^{\min}_Y = \lambda^{\max}_Y = 0$.
    \item Similarly, when $p^{\prime}_{01} = 0$, then $\lambda^{\max}_Y = 0$. $p^{\prime}_{00} - 1$ will at most equal $0$. Therefore $\lambda^{\min}_Y = 0$. So $\lambda^{\min}_Y = \lambda^{\max}_Y = 0$.
    \item When $p^{\prime}_{00} = 1$, $\lambda^{\min}_Y = \max\{p^{\prime}_{01}, 0\}$ and $\lambda^{\max}_Y = \min\{p^{\prime}_{01}, 1\}$. Since $p^{\prime}_{01}$ is a probability and non-negative, $\lambda^{\min}_Y = \lambda^{\max}_Y = p^{\prime}_{01}$.
    \item A similar argument shows that when $p^{\prime}_{01} = 1$, then $\lambda^{\min}_Y = \lambda^{\max}_Y = p^{\prime}_{00}$.
\end{itemize}

Next, we state a paraphrased version of Proposition 4.1 in \cite{gresele2022causal}.
\begin{proposition}
If $\mathcal{C}$ is non-empty, then $\exists \mathbf{c} \in \mathcal{C}$ s.t. $[A\mathbf{c}]_0 = \lambda^{\max}_X$ and $[B\mathbf{c}]_0 = \lambda^{\max}_Y$. 
\end{proposition}

%The next proposition is the contrapositive of the above result, and hence holds as well.
%\begin{proposition}
%If there doesn't exist a $\mathbf{c} \in \mathcal{C}$ such that $[A\mathbf{c}]_0 = \lambda^{\max}_X$ and $[B\mathbf{c}]_0 = \lambda^{\max}_Y$, then $\mathcal{C}$ must be empty. 
%\end{proposition}
%So if we can show that $\not \exists \mathbf{c} \in \mathcal{C}$ such that $[A\mathbf{c}]_0 = \lambda^{\max}_X$ or $[\mathbf{B}\mathbf{c}]_0 = \lambda^{\max}_Y$ , then $\mathcal{C}$ must be empty. 
With these results we are ready to prove Theorem \ref{thm:causal_compatibility}.

%\numair{Statement of theorem 1: When $\mathbb{P}(Z, Y)$ has degenerate conditionals, imposing $[\mathbf{A}\mathbf{c}]_0 = \lambda^{\max}_X$ in combination with the polytope constraints and requiring $\mathbf{c}$ be a valid probability distribution generates necessary and sufficient conditions on $\mathcal{C}$ being non-empty.}
\paragraph{Proof For Theorem \ref{thm:causal_compatibility}} To prove a condition $A$ is necessary and sufficient for $B$, we must prove $A \implies B$ as well as $B \implies A$. In our case, this means we must prove the following two statements:
\begin{itemize}
\item When $\mathcal{C}$ is non-empty, there $\exists \mathbf{c}$ that is a valid probability distribution that satisfies all of the constraints on $\mathcal{C}$ and has $[\mathbf{A}\mathbf{c}]_0 = \lambda^{\max}_X$. 
\item When the conditions derived from requiring $\mathbf{c}$ to satisfy $[\mathbf{A}\mathbf{c}]_0 = \lambda^{\max}_X$ in combination with the polytope constraints are satisfied, then $\mathcal{C}$ is non-empty.
\end{itemize}

Note that Proposition 4.1 from \cite{gresele2022causal} proves the first statement, and for the remaining proof we focus on the second statement. The proof proceeds on a case by case basis for each of the four cases that arise in degenerate conditionals (seen in \ref{degen_cases}), however we only present the proof for $\mathbb{P}(Z = 0 \mid Y = 1) = 0$, and a similar derivation can be performed for the rest.

\subsection{$\mathbb{P}(Z = 0 \mid Y = 1) = 0$}

When $\mathbb{P}(Z = 0 \mid Y = 1) = 0$,  $\lambda^{\max}_Y = \lambda^{\min}_Y = 0$, and the SCM $Y \rightarrow Z$ can be parameterised as below, where where $p^{\prime}_{00}$ denotes $\mathbb{P}(Z = 0 \mid Y = 0)$ and $p^{\prime}_{01} = \mathbb{P}(Z = 0 \mid Y = 1)$. Note that since $\lambda^{\max}_Y = \lambda^{\min}_Y = 0$, all of the inequality constraints on $\mathbf{B}\mathbf{c}$ are now equality constraints. 
\begin{align}
    \mathbf{b}(\lambda_Y) = \begin{pmatrix}0 \\ 1 - p^{\prime}_{00} \\ p^{\prime}_{00} \\ 0\end{pmatrix}
\end{align}

Consequently, the constraint $\mathbf{b}(\lambda_Y) = \mathbf{B}\mathbf{c}$ combined with the nonnegativity of $c_i \in \mathbf{c}$ and $\mathbb{P}(Y = 0)$ and $\mathbb{P}(X = 0)$ being bounded away from $0$ and $1$ enforces the following elements of $\mathbf{c}$ to equal $0$:
\begin{align}
    c_0 = c_1 = c_2 = c_3 = c_4 = c_5 = c_6 = c_7 = c_8 = c_9 = c_{12} = c_{13} = 0
\end{align}

After setting the relevant values of $\mathbf{c}$ to zero, and putting the remaining equality constraints on $\mathbf{c}$ in Reduced Row Echelon Form (RREF) we have the system of equations presented below.

\begin{align}
    c_{10} - c_{15} &= \frac{p_{00} + p_{01} - \mathbb{P}(Y = 0)}{\mathbb{P}(Y = 0)}\\
    c_{11} + c_{15} &= \frac{\mathbb{P}(Y = 0) - p_{00}}{\mathbb{P}(Y = 0)}\\
    c_{14} + c_{15} &= \frac{\mathbb{P}(Y = 0) - p_{01}}{\mathbb{P}(Y = 0)}
\end{align}

Since there are three equations and four unknowns, this system of equations has infinite solutions. However, the additional constraint provided in the form of $[A\mathbf{c}]_0 = \lambda^{\max}_X$ provides a fourth equation that uniquely determines this system of equations, which upon solving yields the following values of $c_i$'s. 
\begin{align}
    [A\mathbf{c}]_0 = \lambda^{\max}_X &\implies \mathbb{P}(Y = 0)c_{10} = \lambda^{\max}_X \implies c_{10} = \frac{\lambda^{\max}_X}{\mathbb{P}(Y = 0)}\\
    c_{15} &= \frac{\lambda^{\max}_X - p_{00} - p_{01} + \mathbb{P}(Y = 0)}{\mathbb{P}(Y = 0)}\\
    c_{11} &= \frac{p_{01} - \lambda^{\max}_X}{\mathbb{P}(Y = 0)}\\
    c_{14} &= \frac{p_{00} - \lambda^{\max}_X}{\mathbb{P}(Y = 0)}
\end{align}
 
Having found the unique value of $\mathbf{c}$ that satisfies the above system of equations, we list the conditions needed for them to be valid probabilities, giving us the following constraints.
    \begin{align}
        0 \leq \lambda^{\max}_X \leq \mathbb{P}(Y = 0)\\
        \lambda^{\max}_X \leq p_{01}\\
        p_{01} - \lambda^{\max}_X \leq \mathbb{P}(Y = 0)
        \lambda^{\max}_X \leq p_{00}\\
        p_{00} - \lambda^{\max}_X \leq \mathbb{P}(Y = 0)\\
        \lambda^{\max}_X \geq p_{00} + p_{01} - \mathbb{P}(Y = 0)\\
        \lambda^{\max}_X \leq p_{00} + p_{01}
    \end{align}

Of these resulting constraints, some are implied by definition, so we discard them, leaving us with the following
\begin{align}
        \lambda^{\max}_X \leq \mathbb{P}(Y = 0)\\
        p_{01} - \lambda^{\max}_X \leq \mathbb{P}(Y = 0)\\
        p_{00} - \lambda^{\max}_X \leq \mathbb{P}(Y = 0)\\
        \lambda^{\max}_X \geq p_{00} + p_{01} - \mathbb{P}(Y = 0)\\
\end{align}

Now we show that when when these conditions holds, the rest of the constraints in $\mathcal{C}$ hold as well, and therefore we have found a point in $\mathcal{C}$, proving it is non-empty. This is demonstrated next by plugging in the values of $c_{10}$, $c_{11}$, $c_{14}$ and $c_{15}$ along with conditions derived above and seeing whether the constraints on $\mathcal{C}$ hold. It should be noted that when $\mathcal{M}_Y$ is a degenerate family, all of the inequality constraints on it become equality constraints that are included in the RREF, so they don't need to be checked. So the only constraints that remain to be checked are the inequality constraints on $\mathbf{A}\mathbf{c}$ are enumerated as
\begin{align}
    \lambda^{\min}_X &\leq c_0 + \mathbb{P}(Y = 1)\{c_1 + c_4 + c_5 \} + \mathbb{P}(Y = 0)\{c_2 + c_8 + c_{10}\}  \leq \lambda^{\max}_X\\
    1 - p_{00} - p_{01} + \lambda^{\min}_X &\leq \mathbb{P}(Y = 0)\{ c_5 + c_7 + c_{13}\} + \mathbb{P}(Y = 1)\{c_{10} + c_{11} + c_{14}\} + c_{15}  \leq 1 - p_{00} - p_{01} + \lambda^{\max}_X\\
    p_{00} - \lambda^{\max}_X &\leq \mathbb{P}(Y = 0)\{ c_4 + c_6 + c_{14}\} + \mathbb{P}(Y = 1)\{c_8 + c_9 + c_{13} \} + c_{12} \leq p_{00} - \lambda^{\min}_X\\
    p_{01} - \lambda^{\max}_X &\leq \mathbb{P}(Y = 0)\{c_1 + c_9 + c_{11}\} + \mathbb{P}(Y = 1)\{ c_2 + c_6 + c_7\} + c_3 \leq p_{01} - \lambda^{\min}_X
\end{align}

Since $c_0 = c_1 = c_2 = c_3 = c_4 = c_5 = c_6 = c_7 = c_8 = c_9 = c_{12} = c_{13}$, each of these simplifies to
\begin{align}
    \lambda^{\min}_X &\leq \mathbb{P}(Y = 0)c_{10}  \leq \lambda^{\max}_X\\
    1 - p_{00} - p_{01} + \lambda^{\min}_X &\leq \mathbb{P}(Y = 1)\{c_{10} + c_{11} + c_{14}\} + c_{15}  \leq 1 - p_{00} - p_{01} + \lambda^{\max}_X\\
    p_{00} - \lambda^{\max}_X &\leq \mathbb{P}(Y = 0)c_{14} \leq p_{00} - \lambda^{\min}_X\\
    p_{01} - \lambda^{\max}_X &\leq \mathbb{P}(Y = 0)c_{11} \leq p_{01} - \lambda^{\min}_X
\end{align}

Inequality I is satisfied by definition. Inequality III and IV are satisfied by definition of $\lambda^{\max}_X \geq \lambda^{\min}_X$. For inequality II, we the result can be seen as follows:
\begin{align}
    \mathbb{P}(Y = 1)\{c_{10} + c_{11} + c_{14}\} + c_{15} = 1 - p_{00} - p_{01} + \lambda^{\max}_X
\end{align}

Which will again satisfy these inequalities. This proves our theorem for the case we present here, and a similar derivation can be repeated for the rest of the cases, i.e. $\mathbb{P}(Z = 0 | Y = 0) = 0$, $\mathbb{P}(Z = 0 | Y = 0) = 1$ and $\mathbb{P}(Z = 0 | Y = 1) = 1$.

\section{Proof For Theorem \ref{thm:pns_bounds}}\label{prf_thm2}

As we have seen in Lemma \ref{lem:link_lambda}, tightening the bounds on PNS is equivalent to tightening the bounds on $\lambda_X$. The bounds on $\lambda_X$ are obtained by solving the linear program with the linear objective function  $[\mathbf{A}\mathbf{c}]_0$ subject to the polytope constraints from Section \ref{convex-polytope}. Each of the bounds is derived on a case-by-case basis for each possible conditionally degenerate distribution $\mathbb{P}(Z, Y)$. Here we denote $\mathbb{P}(Z = 0 \mid Y = 1)$ as $p^{\prime}_{01}$, $\mathbb{P}(Z = 0 \mid Y = 0)$ as $p^{\prime}_{00}$, $\mathbb{P}(Z = 0 \mid X = 1)$ as $p_{01}$ and $\mathbb{P}(Z = 0 \mid X = 0)$ as $p_{00}$ . The bounds are derived by an application of the simplex algorithm, details of which can be found in \cite{dantzig1990origins}.

\subsection{$\mathbb{P}(Z = 0 \mid Y = 1)$}

As we have seen in Section \ref{prf_thm1}, when $\mathbb{P}(Z = 0 \mid Y = 1) = 0$, the equality constraints on $\mathcal{C}$ can be updated as
\begin{align}
    c_{10} - c_{15} &= \frac{p_{00} + p_{01} - \mathbb{P}(Y = 0)}{\mathbb{P}(Y = 0)}\\
    c_{11} + c_{15} &= \frac{\mathbb{P}(Y = 0) - p_{00}}{\mathbb{P}(Y = 0)}\\
    c_{14} + c_{15} &= \frac{\mathbb{P}(Y = 0) - p_{01}}{\mathbb{P}(Y = 0)}
\end{align}

And the following inequality constraints on $\mathbf{A}\mathbf{c}$ still remain, given as
\begin{align}
    \lambda^{\min}_X &\leq \mathbb{P}(Y = 0)c_{10}  \leq \lambda^{\max}_X\\
    1 - p_{00} - p_{01} + \lambda^{\min}_X &\leq \mathbb{P}(Y = 1)\{c_{10} + c_{11} + c_{14}\} + c_{15}  \leq 1 - p_{00} - p_{01} + \lambda^{\max}_X\\
    p_{00} - \lambda^{\max}_X &\leq \mathbb{P}(Y = 0)c_{14} \leq p_{00} - \lambda^{\min}_X\\
    p_{01} - \lambda^{\max}_X &\leq \mathbb{P}(Y = 0)c_{11} \leq p_{01} - \lambda^{\min}_X
\end{align}

To apply the simplex algorithm, we express our minimisation problem as a simplex tableau in the standard form using slack and surplus variables denotes $s_i$ and $e_i$ respectively, where $s_i \geq 0 \quad \forall i$, and similarly for $e_i$. 
{\small
\begin{centering}
\begin{align}
\begin{bmatrix} 
z & c_{10} & c_{11} & c_{14} & c_{15} & s_1 & s_2 & s_3 & s_4 & e_1 & e_2 & e_3 & e_4 & RHS\\
1 & - \mathbb{P}(Y = 0) & 0 & 0 & 0 & 0 & 0 & 0 & 0 & 0 & 0 & 0 & 0 & 0\\
0 & 1 & 0 & 0 & -1 & 0 & 0 & 0 & 0 & 0 & 0 & 0 & 0 & \frac{p_{00} + p_{01} - \mathbb{P}(Y = 0)}{\mathbb{P}(Y = 0)}\\
0 & 0 & 1 & 0 & 1 & 0 & 0 & 0 & 0 & 0 & 0 & 0 & 0 & \frac{\mathbb{P}(Y = 0) - p_{00}}{\mathbb{P}(Y = 0)}\\
0 & 0 & 0 & 1 & 1 & 0 & 0 & 0 & 0 & 0 & 0 & 0 & 0 & \frac{\mathbb{P}(Y = 0) - p_{01}}{\mathbb{P}(Y = 0)}\\
0 & \mathbb{P}(Y = 0) & 0 & 0 & 0 & 1 & 0 & 0 & 0 & 0 & 0 & 0 & 0 & \lambda^{\max}_X\\
0 & \mathbb{P}(Y = 1) & \mathbb{P}(Y = 1) & \mathbb{P}(Y = 1) & 1 & 0 & 1 & 0 & 0 & 0 & 0 & 0 & 0 & 1 - p_{00} - p_{01} + \lambda^{\max}_X\\
0 & 0 & 0 &  \mathbb{P}(Y = 0) &   0 &  0 &  0 &  1 &  0 &   0 &   0 &   0 &   0 & p_{00} - \lambda_X^{\min}\\
0 & 0 &  \mathbb{P}(Y = 0) & 0 & 0 &  0 &  0 &  0 &  1 & 0 & 0 & 0 & 0 & p_{01} - \lambda_X^{\min}\\
0 &  \mathbb{P}(Y = 0) & 0 & 0 & 0 &  0 &  0 &  0 &  0 &  -1 &   0 &   0 &   0 & \lambda_X^{\min}\\
0 & \mathbb{P}(Y = 1) & \mathbb{P}(Y = 1) & \mathbb{P}(Y = 1) & 1 &  0 &  0 &  0 &  0 &   0 &  -1 &   0 &   0 &  \lambda_X^{\min} - p_{00} - p_{01} + 1\\
0 & 0 & 0 &  \mathbb{P}(Y = 0) & 0 & 0 & 0 &  0 &  0 &   0 &   0 &  -1 &   0 & p_{00} - \lambda_X^{\max}\\
0 & 0 &  \mathbb{P}(Y = 0) & 0 & 0 &  0 &  0 &  0 &  0 &   0 &   0 &   0 &  -1 & p_{01} - \lambda_X^{\max}
\end{bmatrix}
\end{align}
\end{centering}
}
In specifying basic and non-basic variables, we must be careful pick a correct combination so we end up with a basic feasible solution. For example, $c_{10}$, $c_{11}$, $c_{14}$ and $c_{15}$ can't all be picked to be non-basic variables because this would lead to a violation of the simplex constraint. 

So we know we must pick slack or surplus variables to be non-basic variables. Based on Theorem \ref{thm:causal_compatibility}, we know that $\lambda^{\max}_X$ will always be a basic feasible solution to this system of equations when this system of equations has a solution. So we proceed to put this system in canonical form with $s_1$ and a combination of other slack and surplus variables as the non-basic variables. The corresponds simplex tableau then becomes
{\small
\begin{align}
\begin{bmatrix}
z & c_{10} & c_{11} & c_{14} & c_{15} & s_1 & s_2 & s_3 & s_4 & e_1 & e_2 & e_3 & e_4 & RHS\\
1 & 0 & 0 & 0 & 0 &  1 & 0 & 0 & 0 &  0 &  0 &  0 &  0 & \lambda_X^{\max}\\
0 & 1 & 0 & 0 & 0 & \frac{1}{\mathbb{P}(Y = 0)} & 0 & 0 & 0 &  0 &  0 &  0 &  0 & \frac{\lambda_X^{\max}}{\mathbb{P}(Y = 0)}\\
0 & 0 & 1 & 0 & 0 &  -\frac{1}{\mathbb{P}(Y = 0)} & 0 & 0 & 0 &  0 &  0 &  0 &  0 & \frac{p_{01} -\lambda_X^{\max}}{\mathbb{P}(Y = 0)}\\
0 & 0 & 0 & 1 & 0 &  -\frac{1}{\mathbb{P}(Y = 0)} & 0 & 0 & 0 &  0 &  0 &  0 &  0 & \frac{p_{00} -\lambda_X^{\max}}{\mathbb{P}(Y = 0)}\\
0 & 0 & 0 & 0 & 1 & \frac{1}{\mathbb{P}(Y = 0)} & 0 & 0 & 0 &  0 &  0 &  0 &  0 & \frac{\lambda_X^{\max} + \mathbb{P}(Y = 0) - p_{00} - p_{01}}{\mathbb{P}(Y = 0)}\\
0 & 0 & 0 & 0 & 0 & -1 & 1 & 0 & 0 &  0 &  0 &  0 &  0 & 0\\
0 & 0 & 0 & 0 & 0 & 1 & 0 & 1 & 0 &  0 &  0 &  0 &  0 & \lambda_X^{\max} -  \lambda_X^{\min}\\
0 & 0 & 0 & 0 & 0 & 1 & 0 & 0 & 1 &  0 &  0 &  0 &  0 & \lambda_X^{\max} -  \lambda_X^{\min}\\
0 & 0 & 0 & 0 & 0 & -1 & 0 & 0 & 0 & -1 &  0 &  0 &  0 & \lambda_X^{\min} -  \lambda_X^{\max}\\
0 & 0 & 0 & 0 & 0 & -1 & 0 & 0 & 0 &  0 & -1 &  0 &  0 & \lambda_X^{\min} -  \lambda_X^{\max}\\
0 & 0 & 0 & 0 & 0 & 1 & 0 & 0 & 0 &  0 &  0 & -1 &  0 & 0\\
0 & 0 & 0 & 0 & 0 & 1 & 0 & 0 & 0 &  0 &  0 &  0 & -1 & 0
\end{bmatrix}
\end{align}
}
However, this tableau has redundancies in the slack and surplus variables which we systematically explore and discard. First, $R_{12}, R_{11}, R_6$ are all in the form $s_1 - s_2 = 0$, which doesn't restrict the range of $s_1$ and is always satisfiable under the assumption of the simplex algorithm. 

Next $R_7, R_8, R_9, R_{10}$ all represent the same constraint, i.e. $s_1 \leq \lambda^{\max}_X - \lambda^{\min}_X$ so we only keep $R7$, which gives the following reduced simplex tableau in canonical form:
\begin{align}
\begin{bmatrix}
z & c_{10} & c_{11} & c_{14} & c_{15} & s_1 & s_3 & RHS\\
1 & 0 & 0 & 0 & 0 &  1 & 0 & \lambda_X^{\max}\\
0 & 1 & 0 & 0 & 0 & \frac{1}{\mathbb{P}(Y = 0)} & 0 & \frac{\lambda_X^{\max}}{\mathbb{P}(Y = 0)}\\
0 & 0 & 1 & 0 & 0 &  -\frac{1}{\mathbb{P}(Y = 0)} & 0 & \frac{p_{01} -\lambda_X^{\max}}{\mathbb{P}(Y = 0)}\\
0 & 0 & 0 & 1 & 0 & -\frac{1}{\mathbb{P}(Y = 0)} & 0 & \frac{p_{00} -\lambda_X^{\max}}{\mathbb{P}(Y = 0)}\\
0 & 0 & 0 & 0 & 1 & \frac{1}{\mathbb{P}(Y = 0)} & 0 & \frac{\lambda_X^{\max} + \mathbb{P}(Y = 0) - p_{00} - p_{01}}{\mathbb{P}(Y = 0)}\\
0 & 0 & 0 & 0 & 0 & 1 & 1 & \lambda_X^{\max} -  \lambda_X^{\min}\\
\end{bmatrix}
\end{align}

Here the non-basic variable is $s_1$, and the basic variables are $z, c_{10}, c_{11}, c_{14}, c_{15}, s_2$, and the basic feasible solution is
\begin{align}
    s_1 &= 0\\
    z &= \lambda_X^{\max}\\
    c_{10} &= \frac{\lambda^{\max}_X}{\mathbb{P}(Y = 0)}\\
    c_{11} &= \frac{p_{01} - \lambda^{\max}_X}{\mathbb{P}(Y = 0)}\\
    c_{14} &= \frac{p_{00} - \lambda^{\max}_X}{\mathbb{P}(Y = 0)}\\
    c_{15} &= \frac{\lambda_X^{\max} + \mathbb{P}(Y = 0) - p_{00} - p_{01}}{\mathbb{P}(Y = 0)}\\
    s_2 &= \lambda^{\max}_X - \lambda^{\min}_X
\end{align}

All the variables can be verified to be positive under the conditions laid out by Theorem \ref{thm:causal_compatibility}. Since the coefficient of the non-basic variable in row 1 is not non-positive, this isn't a minimum. So we use the ratio test to decide which variable to make non-basic.

\begin{itemize}
    \item Ratio test for $R2$: 
    \begin{align}
    c_{10} + \frac{s_1}{\mathbb{P}(Y = 0)} &= \frac{\lambda^{\max}_X}{\mathbb{P}(Y = 0)}\\
    c_{10} &= \frac{\lambda^{\max}_X - s_1}{\mathbb{P}(Y = 0)} \geq 0\\
    s_1 \leq \lambda^{\max}_X
    \end{align}
    \item Ratio test for $R3$
    \begin{align}
        c_{11} - \frac{s_1}{\mathbb{P}(Y = 0)} &= \frac{p_{01} - \lambda_X^{\max}}{\mathbb{P}(Y = 0)}\\
        c_{11} &= p_{01} - \lambda^{\max}_X + s_1 \geq 0\\
        s_1 \geq 0
    \end{align}
    \item Similarly, $R4$ will also impose $s_1 \geq 0$
    \item Ratio test for $R5$
    \begin{align}
        c_{15} + \frac{s_1}{\mathbb{P}(Y = 0)} &= \frac{\lambda^{\max}_X + \mathbb{P}(Y = 0) - p_{00} - p_{01}}{\mathbb{P}(Y = 0)}\\
        c_{15} &= \frac{\lambda^{\max}_X + \mathbb{P}(Y = 0) - p_{00} - p_{01} - s_1}{\mathbb{P}(Y = 0)} \geq 0\\
        s_1 &\leq \lambda^{\max}_X + \mathbb{P}(Y = 0) - p_{00} - p_{01}
    \end{align}
    \item Ratio test for $R6$
    \begin{align}
        s_1 + s_2 &= \lambda^{\max}_X - \lambda^{\min}_X \geq 0\\
        s_1 &\leq \lambda^{\max}_X - \lambda^{\min}_X
    \end{align}
\end{itemize}

The winner of the ratio test and consequently the variable we pivot in will be decided by $\min\{\lambda^{\max}_X, \lambda^{\max}_X + \mathbb{P}(Y = 0) - p_{00} - p_{01}, \lambda^{\max}_X - \lambda^{\min}_X\}$. The minimum here is dependent on the values of $p_{00}$, $p_{01}$ and $\mathbb{P}(Y = 0)$, presented in three cases below

\begin{itemize}
    \item $p_{00} + p_{01} - 1 > 0$: Here $\lambda^{\min}_X = p_{00} + p_{01} - 1$. The minimum will be $\lambda^{\max}_X + \mathbb{P}(Y = 0) - p_{00} - p_{01}$, consequently $c_{15}$ will be come the non-basic variable. 
    
    \item $p_{00} + p_{01} - 1 = 0$: Here $\lambda^{\min}_X = 0$. Then $c_{15}$ will also be the non-basic variable.
    
    \item $p_{00} + p_{01} - 1 < 0$: Here $\lambda^{\min}_X = 0$. This comes down to checking the relationship between $\lambda^{\max}_X$ and $\lambda^{\max}_X + \mathbb{P}(Y = 0) - p_{00} - p_{01}$, analyzed on a case by case basis below.
    \begin{itemize}
        \item $\lambda^{\max}_X > \lambda^{\max}_X + \mathbb{P}(Y = 0) - p_{00} - p_{01}$: This implies $p_{00} + p_{01} - 1 + \mathbb{P}(Y = 1) > 0$, and so $c_{15}$ will be the non-basic variable. 
        \item $\lambda^{\max}_X = \lambda^{\max}_X + \mathbb{P}(Y = 0) - p_{00} - p_{01}$: This implies $p_{00} + p_{01} - 1 + \mathbb{P}(Y = 1) = 0$, and here $c_{10}$ will become the non-basic variable.
        \item $\lambda^{\max}_X < \lambda^{\max}_X + \mathbb{P}(Y = 0) - p_{00} - p_{01}$: This implies $p_{00} + p_{01} - 1 + \mathbb{P}(Y = 1) < 0$, and here $c_{10}$ will become the non-basic variable.
    \end{itemize}
\end{itemize}

So, depending on the situation, we will pivot in $c_{10}$ or $c_{15}$. The simplex tableau after pivoting in $c_{15}$ is given as

\begin{align}
\begin{bmatrix}
    z & c_{10} & c_{11} & c_{14} & c_{15} & s_1 & s_3 & RHS\\
    1 &  0 &  0 &  0 &  -\mathbb{P}(Y = 0) &  0 &  0 & p_{00} + p_{01} - 1 + \mathbb{P}(Y = 1)\\
    0 &  1 &  0 &  0 & -1 & 0 & 0 & \frac{p_{00} + p_{01} -\mathbb{P}(Y = 0)}{\mathbb{P}(Y = 0)}\\
    0 &  0 &  1 &  0 & 1 &  0 &  0 & \frac{\mathbb{P}(Y = 0) - p_{00}}{\mathbb{P}(Y = 0)}\\
    0 &  0 &  0 &  1 & 1 &  0 &  0 & \frac{\mathbb{P}(Y = 0) - p_{01}}{\mathbb{P}(Y = 0)}\\
    0 &  0 &  0 &  0 &  \mathbb{P}(Y = 0) &  1 &  0 &  \lambda^{\max}_X - p_{00} - p_{01} + \mathbb{P}(Y = 0)\\
    0 &  0 &  0 &  0 &  -\mathbb{P}(Y = 0) &  0 &  1 & p_{00} + p_{01} -\mathbb{P}(Y = 0) -\lambda^{\min}_X\\
\end{bmatrix}
\end{align}

And the corresponding basic feasible solution will be
\begin{align}
    c_{15} &= 0\\
    z &= p_{00} + p_{01} - 1 + \mathbb{P}(Y = 1)\\
    c_{10} &= \frac{p_{00} + p_{01} - \mathbb{P}(Y = 0)}{\mathbb{P}(Y = 0)}\\
    c_{11} &= \frac{\mathbb{P}(Y = 0) - p_{00}}{\mathbb{P}(Y = 0)}\\
    c_{14} &= \frac{\mathbb{P}(Y = 0) - p_{01}}{\mathbb{P}(Y = 0)}\\
    s_3 &= p_{00} + p_{01} - \mathbb{P}(Y = 0) - \lambda^{\min}_X
\end{align}

And all the values are positive so this is a basic feasible solution, and this is a minimum because the non-basic variable has a non-positive coefficient. And the corresponding minimum is $p_{00} + p_{01} - 1 + \mathbb{P}(Y = 1)$. 

Similarly, when $c_{10}$ gets pivoted in, the corresponding simplex tableau is
\begin{align}
\begin{bmatrix}
    z & c_{10} & c_{11} & c_{14} & c_{15} & s_1 & s_3 & RHS\\
    1 &  -\mathbb{P}(Y = 0) &  0 &  0 &  0 &  0 &  0 & 0\\
    0 &  \mathbb{P}(Y = 0) &  0 &  0 &  0 &  1 &  0 & \lambda^{\max}_X\\
    0 &      1 &  1 &  0 &  0 &  0 &  0 & \frac{p_{01}}{\mathbb{P}(Y = 0)}\\
    0 &      1 &  0 &  1 &  0 &  0 &  0 & \frac{p_{00}}{\mathbb{P}(Y = 0)}\\
    0 &     -1 &  0 &  0 &  1 &  0 &  0 & \frac{\mathbb{P}(Y = 0) - p_{00} - p_{01}}{\mathbb{P}(Y = 0)}\\
    0 &  -\mathbb{P}(Y = 0) &  0 &  0 &  0 &  0 &  1 &  -\lambda^{\min}_X
\end{bmatrix}
\end{align}

And this case, the basic feasible solution is
\begin{align}
    z &= 0\\
    c_{10} &= 0\\
    s_1 &= \lambda^{\max}_X\\
    c_{11} &= \frac{p_{01}}{\mathbb{P}(Y = 0)}\\
    c_{14} &= \frac{p_{00}}{\mathbb{P}(Y = 0)}\\
    c_{15} &= \frac{\mathbb{P}(Y = 0) - p_{00} - p_{01}}{\mathbb{P}(Y = 0)}\\
    s_2 &= -\lambda^{\min}_X = 0
\end{align}

These two conditions sum up the range restriction $\lambda_X$. Putting it all together, the range restriction of $\lambda_X$ is dependant on the value of $\Phi_X = p_{00} + p_{01} - 1$ as follows:
\begin{itemize}
    \item $\Phi_X \geq 0$: The updated range of $\lambda_X$ will be $[\Phi_X + \mathbb{P}(Y = 1), \lambda^{\max}_X]$.
    \item $\Phi_X < 0$: This is split into two cases, where
    \begin{itemize}
        \item If $\Phi_X + \mathbb{P}(Y = 1) \leq 0$, then the updated range of $\lambda_X$ will be $[\lambda^{\min}_X, \lambda^{\max}_X]$. 
        \item $\Phi_X + \mathbb{P}(Y = 1) > 0$: The updated range of $\lambda^{\max}_X$ is $[\Phi_X + \mathbb{P}(Y = 1), \lambda^{\max}_X]$
    \end{itemize}
\end{itemize}

A similar derivation provides equivalent results on range restriction on $\lambda_X$ for each of the cases when $\mathbb{P}(Z, Y)$ has degenerate conditionals. We leave the derivation as an exercise to the reader and only provide the results below.

\subsection{Statement of theorem for $p^{\prime}_{00} = 0$}

When $\lambda^{\max}_Y = \lambda^{\min}_Y = p^{\prime}_{00} = 0$ and the conditions outlined in distribution compatibility hold, the enforcement of counterfactual consistency on the range of $\lambda_X$ is dependant on value of the normalized effect strength $\Phi_X$ in the following way:
\begin{itemize}
    \item $\Phi_X \geq 0$: The updated range of $\lambda_X$ will be $[\Phi_X + \mathbb{P}(Y = 0), \lambda^{\max}_X]$
    \item $\Phi_X < 0$: This is split into two cases, where
    \begin{itemize}
        \item $\Phi_X + \mathbb{P}(Y = 0) \leq 0$: Then there will be no falsification, and the range of $\lambda_X$ will be $[0, \lambda^{\max}_X]$.
        \item $\Phi_X + \mathbb{P}(Y = 0) > 0$: The updated range of $\lambda_X$ will be $[\Phi_X + \mathbb{P}(Y = 0), \lambda^{\max}_X]$.
    \end{itemize}
\end{itemize}

\subsection{Statement of theorem for $p^{\prime}_{00} = 1$}

The result is split into the following cases
\begin{itemize}
    \item $\Phi_X \leq 0$: Then the updates range of $\lambda_X$ is $[\mathbb{P}(Y = 0), \lambda^{\max}_X]$.
    \item $\Phi_X > 0$: This is split into two cases
    \begin{itemize}
        \item $\Phi_X - \mathbb{P}(Y = 0) \geq 0$: Then no falsification and the range of $\lambda_X$ is $[\Phi_X, \lambda^{\max}_X]$.
        \item $\Phi_X - \mathbb{P}(Y = 0) < 0$: The updated range of $\lambda_X$ will be $[\mathbb{P}(Y = 0), \lambda^{\max}_X]$
    \end{itemize}
\end{itemize}

\subsection{Statement of theorem for $p^{\prime}_{01} = 1$}

The result is similarly split into the follow cases
\begin{itemize}
    \item $\Phi_X \leq 0$: Then the updates range of $\lambda_X$ is $[\mathbb{P}(Y = 1), \lambda^{\max}_X]$.
    \item $\Phi_X > 0$: This is split into two cases
    \begin{itemize}
        \item $\Phi_X - \mathbb{P}(Y = 1) \geq 0$: Then no falsification and the range of $\lambda_X$ is $[\Phi_X, \lambda^{\max}_X]$.
        \item $\Phi_X - \mathbb{P}(Y = 1) < 0$: The updated range of $\lambda_X$ will be $[\mathbb{P}(Y = 1), \lambda^{\max}_X]$
    \end{itemize}
\end{itemize}

\subsection{Restatement of Bounds on $\lambda_X$}
When $\mathbb{P}(Z, Y)$ has degenerate conditionals, all four of the above results can be succinctly stated using indicator functions as follows the updated lower bound on $\lambda_X$ can be stated as
\begin{align}
\mathbb{D}_0 &\equiv \mathbb{I}(p^{\prime}_{00} = 0)\mathbb{P}(Y = 0) + \mathbb{I}(p^{\prime}_{01} = 0)\mathbb{P}(Y = 1)\\
\mathbb{D}_1 &\equiv \mathbb{I}(p^{\prime}_{00} = 1)\mathbb{P}(Y = 0) + \mathbb{I}(p^{\prime}_{01} = 1)\mathbb{P})(Y = 1)\\
& \quad \max\{\Phi_X + \mathbb{D}_0, \mathbb{D}_1\} \leq \lambda_X 
\end{align}

\subsection{Updated Bounds on PNS}

Since PNS is given as $p_{00} - \lambda_X$, these bounds can be straightforwardly translated. As the upper bound on $\lambda_X$ remains unchanged, so does the lower bound on PNS. However the upper bound on PNS will decrease, and we present the updated bounds on PNS below.
\begin{align}
\max[0, p_{11} - p_{10}] \leq PNS \leq \min\{p_{00} - \mathbb{D}_1, p_{11} - \mathbb{D}_0\}
\end{align}

\section{Extension To Additional Counterfactual Probabilities}\label{additional_counterfactuals}

While we limit our focus to PNS, our falsification of SCMs corresponding to certain values of $\lambda_X$ can also used to bound other counterfactuals, an example of which we discuss now. We demonstrate this with two more counterfactuals of interest, which inspired by PN and PS and are named similarly
\begin{itemize}
    \item Probability of sufficient non-monotonicity
    \begin{align}
        \mathbb{P}(Z = 0 \mid X = 0, Z = 1, do(X = 1))
    \end{align}
    \item Probability of necessary non-monotonicity
    \begin{align}
        \mathbb{P}(Z = 1 \mid X = 1, Z = 0, do(X = 0))
    \end{align}
\end{itemize}

\subsubsection{Probability of sufficient non-monotonicity}

This can be calculated using Pearl's three-step process of abduction, action and prediction \cite{pearl2009causality}. We start by computing $\mathbb{P}(N_Z \mid X = 0, Z = 1)$ using Bayes Theorem
\begin{align}
    \mathbb{P}(N_Z \mid X = 0, Z = 1) &= \frac{\mathbb{P}(X = 0, Z = 1 \mid N_Z)\mathbb{P}(N_Z)}{\mathbb{P}(X = 0, Z = 1)}\\
    &= \frac{\mathbb{P}(Z = 1 \mid X = 0, N_Z)\mathbb{P}(X = 0 \mid N_Z)\mathbb{P}(N_Z)}{\mathbb{P}(X = 0, Z = 1)}\\
    &= \frac{\mathbb{P}(Z = 1 \mid X = 0, N_Z)\mathbb{P}(N_Z)}{\mathbb{P}(Z = 1 \mid X = 0)}\\
\end{align}

And so $\mathbb{P}(N_Z \mid X = 0, Z = 1) = [0, \frac{1 - p_{00} - p_{01} - \lambda_X}{\mathbb{P}(Z = 1 \mid X = 0)}, 0, \frac{p_{01} - \lambda_X}{\mathbb{P}(Z = 1 \mid X = 0)}]$. Consequently, the value of the counterfactual probability is given as
\begin{align}
    \mathbb{P}(Z = 0 \mid X = 0, Z = 1, do(X = 1)) = \frac{p_{01} - \lambda_X}{\mathbb{P}(Z = 1 \mid X = 0)}
\end{align}

\subsubsection{Probability of necessary non-monotonicity}

Following a similar procedure, we get
\begin{align}
    \mathbb{P}(Z = 1 \mid X = 1, Z = 0, do(X = 0)) = \frac{p_{01} - \lambda_X}{\mathbb{P}(Z = 0 \mid X = 1)}
\end{align}

So we see that other counterfactuals can also be bounded through restricting the range of $\lambda_X$.

\newpage
\bibliography{references}
\end{document}